\documentclass[twoside,english,thmsa]{article}
\usepackage[T1]{fontenc}
\usepackage[latin9]{inputenc}
\usepackage{babel}

\usepackage{amstext}
\usepackage{graphicx}
\usepackage{esint}
\usepackage[unicode=true, pdfusetitle,
 bookmarks=true,bookmarksnumbered=false,bookmarksopen=false,
 breaklinks=false,pdfborder={0 0 1},backref=false,colorlinks=false]
 {hyperref}
\begin{document}

\title{A Self-Organising Neural Network for Processing Data from Multiple
Sensors%
\thanks{This unpublished draft paper accompanied a talk that was given at
the Conference on Neural Networks for Computing, 4-7 April 1995, Snowbird.%
}}

\author{S P Luttrell}
\maketitle
\begin{abstract}
This paper shows how a folded Markov chain network can be applied
to the problem of processing data from multiple sensors, with an emphasis
on the special case of 2 sensors. It is necessary to design the network
so that it can transform a high dimensional input vector into a posterior
probability, for which purpose the partitioned mixture distribution
network is ideally suited. The underlying theory is presented in detail,
and a simple numerical simulation is given that shows the emergence
of ocular dominance stripes.
\end{abstract}

\section{Theory}

\label{Sect:Theory}

\subsection{Neural Network Model}

\label{SubSect:NNInterp}

In order to fix ideas, it is useful to give an explicit {}``neural
network'' interpretation to the theory that will be developed. The
model will consist of 2 layers of nodes. The input layer has a {}``pattern
of activity'' that represents the components of the input vector
$\mathbf{x}$, and the output layer has a pattern of activity that
is the collection of activities of each output node. The activities
in the output layer depend only on the activities in the input layer.
If an input vector $\mathbf{x}$ is presented to this network, then
each output node {}``fires'' discretely at a rate that corresponds
to its activity. After $n$ nodes have fired the probabilistic description
of the relationship between the input and output of the network is
given by $\Pr\left(\mathbf{y}_{1},\mathbf{y}_{2},\cdots,\mathbf{y}_{n}|\mathbf{x}\right)$,
where $\mathbf{y}_{i}$ is the location in the output layer (assumed
to be on a rectangular lattice of size $\mathbf{m}$)\ of the $i^{th}$
node that fires. In this paper it will be assumed that the order in
which the $n$ nodes fire is not observed, in which case $\Pr\left(\mathbf{y}_{1},\mathbf{y}_{2},\cdots,\mathbf{y}_{n}|\mathbf{x}\right)$
is a sum of probabilities over all $n!$ permutations of $\left(\mathbf{y}_{1},\mathbf{y}_{2},\cdots,\mathbf{y}_{n}\right)$,
which is a symmetric function of the $\mathbf{y}_{i}$, by construction.

The theory that is introduced in section \ref{SubSect:ProbEncodeDecode}
concerns the special case $n=1$. In the $n=1$ case the probabilistic
description $\Pr\left(\mathbf{y|x}\right)$ is proportional to the
firing rate of node $\mathbf{y}$ in response to input $\mathbf{x}$.
When $n>1$ there is an indirect relationship between the probabilistic
description $\Pr\left(\mathbf{y}_{1},\mathbf{y}_{2},\cdots,\mathbf{y}_{n}|\mathbf{x}\right)$
and the firing rate of node $\mathbf{y}$, which is given by the marginal
probability \begin{equation}
\Pr\left(\mathbf{y|x}\right)=\sum_{\mathbf{y}_{2},\cdots,\mathbf{y}_{n}}^{\mathbf{m}}\Pr\left(\mathbf{y},\mathbf{y}_{2},\cdots,\mathbf{y}_{n}|\mathbf{x}\right)\end{equation}
It is important to maintain this distinction between events that are
observed (i.e. $\left(\mathbf{y}_{1},\mathbf{y}_{2},\cdots,\mathbf{y}_{n}\right)$
given $\mathbf{x}$) and the probabilistic description of the events
that are observed (i.e. $\Pr\left(\mathbf{y}_{1},\mathbf{y}_{2},\cdots,\mathbf{y}_{n}|\mathbf{x}\right)$).
The only possible exception is in the $n\rightarrow\infty$ limit,
where $\Pr\left(\mathbf{y}_{1},\mathbf{y}_{2},\cdots,\mathbf{y}_{n}|\mathbf{x}\right)$
has all of its probability concentrated in the vicinity of those $\left(\mathbf{y}_{1},\mathbf{y}_{2},\cdots,\mathbf{y}_{n}\right)$
that are consistent with the observed long-term average firing rate
of each node. It is essential to consider the $n>1$ case to obtain
the results that are described in this paper.

\subsection{Probabilistic Encoder/Decoder}

\label{SubSect:ProbEncodeDecode}

A theory of self-organising networks based on an analysis of a probabilistic
encoder/decoder was presented in \cite{Ref:LuttrellSOM}. It deals
with the $n=1$ case referred to in section \ref{SubSect:NNInterp}.
The objective function that needs to be minimised in order to optimise
a network in this theory is the Euclidean distortion $D$ defined
as \begin{equation}
D\equiv\sum_{\mathbf{y}=\mathbf{1}}^{\mathbf{m}}\int d\mathbf{x}\, d\mathbf{x}^{\prime}\Pr\left(\mathbf{x}\right)\Pr\left(\mathbf{y}|\mathbf{x}\right)\Pr\left(\mathbf{x}^{\prime}\mathbf{|y}\right)\left\Vert \mathbf{x}-\mathbf{x}^{\prime}\right\Vert ^{2}\label{Eq:Objective}\end{equation}
where $\mathbf{x}$ is an input vector, $\mathbf{y}$ is a coded version
of $\mathbf{x}$ (a vector index on a $d$-dimensional rectangular
lattice of size $\mathbf{m}$), $\mathbf{x}^{\prime}$ is a reconstructed
version of $\mathbf{x}$ from $\mathbf{y}$, $\Pr\left(\mathbf{x}\right)$
is the probability density of input vectors, $\Pr\left(\mathbf{y}|\mathbf{x}\right)$
is a probabilistic encoder, and $\Pr\left(\mathbf{x}^{\prime}\mathbf{|y}\right)$
is a probabilistic decoder which is specified by Bayes' theorem as
\begin{equation}
\Pr\left(\mathbf{x|y}\right)=\frac{\Pr\left(\mathbf{y}|\mathbf{x}\right)\Pr\left(\mathbf{x}\right)}{\int d\mathbf{x}^{\prime}\Pr\left(\mathbf{y}|\mathbf{x}^{\prime}\right)\Pr\left(\mathbf{x}^{\prime}\right)}\label{Eq:Bayes}\end{equation}
$D$ can be rearranged into the form \cite{Ref:LuttrellSOM} \begin{equation}
D=2\sum_{\mathbf{y}=\mathbf{1}}^{\mathbf{m}}\int d\mathbf{x}\Pr\left(\mathbf{x}\right)\Pr\left(\mathbf{y}|\mathbf{x}\right)\left\Vert \mathbf{x}-\mathbf{x}^{\prime}\left(\mathbf{y}\right)\right\Vert ^{2}\label{Eq:ObjectiveSimplified}\end{equation}
where the reference vectors $\mathbf{x}^{\prime}\left(\mathbf{y}\right)$
are defined as \begin{equation}
\mathbf{x}^{\prime}\left(\mathbf{y}\right)\equiv\int d\mathbf{x}\Pr\left(\mathbf{x|y}\right)\,\mathbf{x}\label{Eq:RefVect}\end{equation}
Although equation \ref{Eq:Objective} is symmetric with respect to
interchanging the encoder and decoder, equation \ref{Eq:ObjectiveSimplified}
is not. This is because Bayes' theorem has made explicit the dependence
of $\Pr\left(\mathbf{x|y}\right)$ on $\Pr\left(\mathbf{y}|\mathbf{x}\right)$.
From a neural network viewpoint $\Pr\left(\mathbf{y}|\mathbf{x}\right)$
describes the feed-forward transformation from the input layer to
the output layer, and $\mathbf{x}^{\prime}\left(\mathbf{y}\right)$
describes the feed-back transformation that is implied from the output
layer to the input layer. The feed-back transformation is necessary
to implement the objective function that has been chosen here.

Minimisation of $D$ with respect to all free parameters leads to
an optimal encoder/decoder. In equation \ref{Eq:ObjectiveSimplified}
the $\Pr\left(\mathbf{y}|\mathbf{x}\right)$ are the only free parameters,
because $\mathbf{x}^{\prime}\left(\mathbf{y}\right)$ is fixed by
equation \ref{Eq:RefVect}. However, in practice, both $\Pr\left(\mathbf{y}|\mathbf{x}\right)$
and $\mathbf{x}^{\prime}\left(\mathbf{y}\right)$ may be treated as
free parameters \cite{Ref:LuttrellSOM}, because $\mathbf{x}^{\prime}\left(\mathbf{y}\right)$
satisfy equation \ref{Eq:RefVect} at stationary points of $D$ with
respect to variation of $\mathbf{x}^{\prime}\left(\mathbf{y}\right)$.

\subsection{Posterior Probability Model}

\label{SubSect:PostProb}

The probabilistic encoder/decoder requires an explicit functional
form for the posterior probability $\Pr\left(\mathbf{y}|\mathbf{x}\right)$.
A convenient expression is \begin{equation}
\Pr\left(\mathbf{y}|\mathbf{x}\right)=\frac{Q\left(\mathbf{x}|\mathbf{y}\right)}{\sum_{\mathbf{y}^{\prime}=\mathbf{1}}^{\mathbf{m}}Q\left(\mathbf{x}|\mathbf{y}^{\prime}\right)}\label{Eq:PostProb}\end{equation}
where $Q\left(\mathbf{x}|\mathbf{y}\right)>0$ can be regarded as
a node {}``activity'', and $\sum_{\mathbf{y}=\mathbf{1}}^{\mathbf{m}}P\left(\mathbf{y}|\mathbf{x}\right)=1$.
Any non-negative function can be used for $Q\left(\mathbf{x}|\mathbf{y}\right)$,
such as a sigmoid (which satisfies $0\leq Q\left(\mathbf{x}|\mathbf{y}\right)\leq1$)
\begin{equation}
Q\left(\mathbf{x}|\mathbf{y}\right)=\frac{1}{1+\exp\left(-\mathbf{w}\left(\mathbf{y}\right)\cdot\mathbf{x}-b\left(\mathbf{y}\right)\right)}\label{Eq:Sigmoid}\end{equation}
where $\mathbf{w}\left(\mathbf{y}\right)$ and $b\left(\mathbf{y}\right)$
are a weight vector and bias, respectively.

A drawback to the use of equation \ref{Eq:PostProb} is that it does
not permit it to scale well to input vectors that have a large dimensionality.
This problem arises from the restricted functional form allowed for
$Q\left(\mathbf{x}|\mathbf{y}\right)$. A solution was presented in
\cite{Ref:LuttrellPMD} \begin{equation}
\Pr\left(\mathbf{y}|\mathbf{x}\right)=\frac{1}{M}\, Q\left(\mathbf{x}|\mathbf{y}\right)\sum_{\mathbf{y}^{\prime}\in\tilde{N}\left(\mathbf{y}\right)}\frac{1}{\sum_{\mathbf{y}^{\prime\prime}\in N\left(\mathbf{y}^{\prime}\right)}Q\left(\mathbf{x}|\mathbf{y}^{\prime\prime}\right)}\label{Eq:PostProbPMD}\end{equation}
where $M\equiv m_{1}m_{2}\cdots m_{d}$, and $N\left(\mathbf{y}\right)$
is a set of lattice points that are deemed to be {}``in the neighbourhood
of'' the lattice point $\mathbf{y}$, and $\tilde{N}\left(\mathbf{y}\right)$
is the inverse neighbourhood defined as the set of lattice points
that have lattice point $\mathbf{y}$ in their neighbourhood. This
expression for $\Pr\left(\mathbf{y}|\mathbf{x}\right)$ satisfies
$\sum_{\mathbf{y}=1}^{\mathbf{m}}P\left(\mathbf{y}|\mathbf{x}\right)=1$
(see appendix \ref{App:PostProbPMDNorm}). It is convenient to define
\begin{equation}
\Pr\left(\mathbf{y}|\mathbf{x};\mathbf{y}^{\prime}\right)\equiv\frac{Q\left(\mathbf{x}|\mathbf{y}\right)}{\sum_{\mathbf{y}^{\prime\prime}\in N\left(\mathbf{y}^{\prime}\right)}Q\left(\mathbf{x}|\mathbf{y}^{\prime\prime}\right)}\label{Eq:PostProbMD}\end{equation}
which is another posterior probability, by construction. It includes
the effect of the output nodes that are in the neighbourhood of node
$\mathbf{y}^{\prime}$ only. $\Pr\left(\mathbf{y}|\mathbf{x};\mathbf{y}^{\prime}\right)$
is thus a localised posterior probability derived from a localised
subset of the node activities. This allows equation \ref{Eq:PostProbPMD}
to be written as $\Pr\left(\mathbf{y}|\mathbf{x}\right)=\frac{1}{M}\sum_{\mathbf{y}^{\prime}\in\tilde{N}\left(\mathbf{y}\right)}\Pr\left(\mathbf{y}|\mathbf{x};\mathbf{y}^{\prime}\right)$,
so $\Pr\left(\mathbf{y}|\mathbf{x}\right)$ is the average of the
posterior probabilities at node $\mathbf{y}$ arising from each of
the localised subsets that happens to include node $\mathbf{y}$.

\subsection{Multiple Firing Model}

\label{SubSect:MultiFire}

The model may be extended to the case where $n$ output nodes fire.
$\Pr\left(\mathbf{y}|\mathbf{x}\right)$ is then replaced by $\Pr\left(\mathbf{y}_{1},\mathbf{y}_{2},\cdots\mathbf{y}_{n}|\mathbf{x}\right)$,
which is the probability that $\left(\mathbf{y}_{1},\mathbf{y}_{2},\cdots\mathbf{y}_{n}\right)$
are the first $n$ nodes to fire (in that order). With this modification,
$D$ becomes \begin{equation}
D=2\sum_{\mathbf{y}_{1},\mathbf{y}_{2},\cdots\mathbf{y}_{n}=\mathbf{1}}^{\mathbf{m}}\int d\mathbf{x}\Pr\left(\mathbf{x}\right)\Pr\left(\mathbf{y}_{1},\mathbf{y}_{2},\cdots\mathbf{y}_{n}|\mathbf{x}\right)\left\Vert \mathbf{x}-\mathbf{x}^{\prime}\left(\mathbf{y}_{1},\mathbf{y}_{2},\cdots\mathbf{y}_{n}\right)\right\Vert ^{2}\label{Eq:ObjectiveMultiFire}\end{equation}
where the reference vectors $\mathbf{x}^{\prime}\left(\mathbf{y}_{1},\mathbf{y}_{2},\cdots\mathbf{y}_{n}\right)$
are defined as \begin{equation}
\mathbf{x}^{\prime}\left(\mathbf{y}_{1},\mathbf{y}_{2},\cdots\mathbf{y}_{n}\right)\equiv\int d\mathbf{x}\Pr\left(\mathbf{x|y}_{1},\mathbf{y}_{2},\cdots\mathbf{y}_{n}\right)\,\mathbf{x}\label{Eq:RefVectMultiFire}\end{equation}
The dependence of $\Pr\left(\mathbf{y}_{1},\mathbf{y}_{2},\cdots\mathbf{y}_{n}|\mathbf{x}\right)$
and $\mathbf{x}^{\prime}\left(\mathbf{y}_{1},\mathbf{y}_{2},\cdots\mathbf{y}_{n}\right)$
on $n$ output node locations complicates this result. Assume that
$\Pr\left(\mathbf{y}_{1},\mathbf{y}_{2},\cdots\mathbf{y}_{n}|\mathbf{x}\right)$
is a symmetric function of its $\left(\mathbf{y}_{1},\mathbf{y}_{2},\cdots\mathbf{y}_{n}\right)$
arguments, which corresponds to ignoring the order in which the first
$n$ nodes choose to fire (i.e. $\Pr\left(\mathbf{y}_{1},\mathbf{y}_{2},\cdots\mathbf{y}_{n}|\mathbf{x}\right)$
is a sum over all permutations of $\left(\mathbf{y}_{1},\mathbf{y}_{2},\cdots\mathbf{y}_{n}\right)$).
For simplicity, assume that the nodes fire independently so that $\Pr\left(\mathbf{y}_{1},\mathbf{y}_{2}|\mathbf{x}\right)=\Pr\left(\mathbf{y}_{1}|\mathbf{x}\right)\Pr\left(\mathbf{y}_{2}|\mathbf{x}\right)$
(see appendix \ref{App:UpperBound} for the general case where $\Pr\left(\mathbf{y}_{1},\mathbf{y}_{2}|\mathbf{x}\right)$
does not factorise). $D$ may be shown to satisfy the inequality $D\leq D_{1}+D_{2}$
(see appendix \ref{App:UpperBound}), where \begin{eqnarray}
D_{1} & \equiv & \frac{2}{n}\sum_{\mathbf{y}=1}^{\mathbf{m}}\int d\mathbf{x}\Pr\left(\mathbf{x}\right)\Pr\left(\mathbf{y}|\mathbf{x}\right)\left\Vert \mathbf{x}-\mathbf{x}^{\prime}\left(\mathbf{y}\right)\right\Vert ^{2}\nonumber \\
D_{2} & \equiv & \frac{2\left(n-1\right)}{n}\int d\mathbf{x}\Pr\left(\mathbf{x}\right)\left\Vert \sum_{\mathbf{y}=\mathbf{1}}^{\mathbf{m}}\Pr\left(\mathbf{y}|\mathbf{x}\right)\left(\mathbf{x}-\mathbf{x}^{\prime}\left(\mathbf{y}\right)\right)\right\Vert ^{2}\label{Eq:UpperBoundPieces}\end{eqnarray}
$D_{1}$ and $D_{2}$ are both non-negative. $D_{1}\rightarrow0$
as $n\rightarrow\infty$, and $D_{2}=0$ when $n=0$, so the $D_{1}$
term is the sole contribution to the upper bound when $n=0$, and
the $D_{2}$ term provides the dominant contribution as $n\rightarrow\infty$.
The difference between the $D_{1}$ and the $D_{2}$ terms is the
location of the $\sum_{\mathbf{y}=\mathbf{1}}^{\mathbf{m}}\Pr\left(\mathbf{y}|\mathbf{x}\right)\left(\cdots\right)$
average: in the $D_{2}$ term it averages a vector quantity, whereas
in the $D_{1}$ term it averages a Euclidean distance. The $D_{2}$
term will therefore exhibit interference effects, whereas the $D_{1}$
term will not.

\subsection{Probability Leakage}

\label{SubSect:Leakage}

The model may be further extended to the case where the probability
that a node fires is a weighted average of the underlying probabilities
that the nodes in its vicinity fire. Thus $\Pr\left(\mathbf{y}|\mathbf{x}\right)$
becomes \begin{equation}
\Pr\left(\mathbf{y}|\mathbf{x}\right)\rightarrow\sum_{\mathbf{y}^{\prime}=\mathbf{1}}^{\mathbf{m}}\Pr\left(\mathbf{y}|\mathbf{y}^{\prime}\right)\Pr\left(\mathbf{y}^{\prime}|\mathbf{x}\right)\label{Eq:Leakage}\end{equation}
where $\Pr\left(\mathbf{y}|\mathbf{y}^{\prime}\right)$ is the conditional
probability that node $\mathbf{y}$ fires given that node $\mathbf{y}^{\prime}$
would have liked to fire. In a sense, $\Pr\left(\mathbf{y}|\mathbf{y}^{\prime}\right)$
describes a {}``leakage'' of probability from node $\mathbf{y}^{\prime}$
that onto node $\mathbf{y}$. $\Pr\left(\mathbf{y}|\mathbf{y}^{\prime}\right)$
then plays the role of a soft {}``neighbourhood function'' for node
$\mathbf{y}^{\prime}$. This expression for $\Pr\left(\mathbf{y}|\mathbf{x}\right)$
can be used wherever a plain $\Pr\left(\mathbf{y}|\mathbf{x}\right)$
has been used before. The main purpose of introducing leakage is to
encourage neighbouring nodes to perform a similar function. This occurs
because the effect of leakage is to soften the posterior probability
$\Pr\left(\mathbf{y}|\mathbf{x}\right)$, and thus reduce the ability
to reconstruct $\mathbf{x}$ accurately from knowledge of $\mathbf{y}$,
which thus increases the average Euclidean distortion $D$. To reduce
the damage that leakage causes, the optimisation must ensure that
nodes that leak probability onto each other have similar properties,
so that it does not matter much that they leak.

\subsection{The Model}

\label{SubSect:Model}

The focus of this paper is on minimisation of the upper bound $D_{1}+D_{2}$
(see equation \ref{Eq:UpperBoundPieces})\ to $D$ in the multiple
firing model, using a scalable posterior probability $\Pr\left(\mathbf{y}|\mathbf{x}\right)$
(see equation \ref{Eq:PostProbPMD}), with the effect of activity
leakage $\Pr\left(\mathbf{y}|\mathbf{y}^{\prime}\right)$ taken into
account (see equation \ref{Eq:Leakage}). Gathering all of these pieces
together yields \begin{eqnarray}
D_{1} & = & \frac{2}{n\, M}\int d\mathbf{x}\Pr\left(\mathbf{x}\right)\sum_{\mathbf{y}=\mathbf{1}}^{\mathbf{m}}\sum_{\mathbf{y}^{\prime}=\mathbf{1}}^{\mathbf{m}}\Pr\left(\mathbf{y}|\mathbf{y}^{\prime}\right)\sum_{\mathbf{y}^{\prime\prime}\in\tilde{N}\left(\mathbf{y}^{\prime}\right)}\Pr\left(\mathbf{y}^{\prime}|\mathbf{x};\mathbf{y}^{\prime\prime}\right)\left\Vert \mathbf{x}-\mathbf{x}^{\prime}\left(\mathbf{y}\right)\right\Vert ^{2}\nonumber \\
D_{2} & = & \frac{2\left(n-1\right)}{n\, M^{2}}\int d\mathbf{x}\Pr\left(\mathbf{x}\right)\label{Eq:ObjectiveModel}\\
 &  & \times\left\Vert \sum_{\mathbf{y}=\mathbf{1}}^{\mathbf{m}}\sum_{\mathbf{y}^{\prime}=\mathbf{1}}^{\mathbf{m}}\Pr\left(\mathbf{y}|\mathbf{y}^{\prime}\right)\sum_{\mathbf{y}^{\prime\prime}\in\tilde{N}\left(\mathbf{y}^{\prime}\right)}\Pr\left(\mathbf{y}^{\prime}|\mathbf{x};\mathbf{y}^{\prime\prime}\right)\left(\mathbf{x}-\mathbf{x}^{\prime}\left(\mathbf{y}\right)\right)\right\Vert ^{2}\nonumber \end{eqnarray}
where $\Pr\left(\mathbf{y}|\mathbf{x};\mathbf{y}^{\prime}\right)\equiv\frac{Q\left(\mathbf{x}|\mathbf{y}\right)}{\sum_{\mathbf{y}^{\prime\prime}\in N\left(\mathbf{y}^{\prime}\right)}Q\left(\mathbf{x}|\mathbf{y}^{\prime\prime}\right)}$.

In order to ensure that the model is truly scalable, it is necessary
to restrict the dimensionality of the reference vectors. In equation
\ref{Eq:ObjectiveModel} $\dim\mathbf{x}^{\prime}\left(\mathbf{y}\right)=\dim\mathbf{x}$,
which is not acceptable in a scalable network. In practice, it will
be assumed any properties of node $\mathbf{y}$ that are vectors in
input space will be limited to occupy an {}``input window'' of restricted
size that is centred on node $\mathbf{y}$. This restriction applies
to the node reference vector $\mathbf{x}^{\prime}\left(\mathbf{y}\right)$,
which prevents $D_{1}+D_{2}$ from being fully minimised, because
$\mathbf{x}^{\prime}\left(\mathbf{y}\right)$ is allowed to move only
in a subspace of the full-dimensional input space. However, useful
results can nevertheless be obtained, so this restriction is acceptable.

\subsection{Optimisation}

\label{SubSect:Optimise}

Optimisation is achieved by minimising $D_{1}+D_{2}$ with respect
to its free parameters. Thus the derivatives with respect to $\mathbf{x}^{\prime}\left(\mathbf{y}\right)$
are given by \begin{eqnarray}
\frac{\partial D_{1}}{\partial\mathbf{x}^{\prime}\left(\mathbf{y}\right)} & = & -\frac{4}{n\, M}\int d\mathbf{x}\Pr\left(\mathbf{x}\right)\,\mathbf{f}_{1}\left(\mathbf{x},\mathbf{y}\right)\nonumber \\
\frac{\partial D_{2}}{\partial\mathbf{x}^{\prime}\left(\mathbf{y}\right)} & = & -\frac{4\left(n-1\right)}{n\, M^{2}}\int d\mathbf{x}\Pr\left(\mathbf{x}\right)\,\mathbf{f}_{2}\left(\mathbf{x},\mathbf{y}\right)\label{Eq:DerivRefVect}\end{eqnarray}
and the variations with respect to $Q\left(\mathbf{x}|\mathbf{y}\right)$
are given by \begin{eqnarray}
\delta D_{1} & = & \frac{2}{n\, M}\sum_{\mathbf{y}=\mathbf{1}}^{\mathbf{m}}\int d\mathbf{x}\Pr\left(\mathbf{x}\right)\, g_{1}\left(\mathbf{x},\mathbf{y}\right)\,\delta\log Q\left(\mathbf{x}|\mathbf{y}\right)\nonumber \\
\delta D_{2} & = & \frac{4\left(n-1\right)}{n\, M^{2}}\sum_{\mathbf{y}=\mathbf{1}}^{\mathbf{m}}\int d\mathbf{x}\Pr\left(\mathbf{x}\right)\, g_{2}\left(\mathbf{x},\mathbf{y}\right)\,\delta\log Q\left(\mathbf{x}|\mathbf{y}\right)\label{Eq:VaryProb}\end{eqnarray}
The functions $\mathbf{f}_{1}\left(\mathbf{x},\mathbf{y}\right)$,
$\,\mathbf{f}_{2}\left(\mathbf{x},\mathbf{y}\right)$, $g_{1}\left(\mathbf{x},\mathbf{y}\right)$,
and $g_{2}\left(\mathbf{x},\mathbf{y}\right)$ are derived in appendix
\ref{App:Derivatives}. Inserting a sigmoidal function $Q\left(\mathbf{x}|\mathbf{y}\right)=\frac{1}{1+\exp\left(-\mathbf{w}\left(\mathbf{y}\right)\cdot\mathbf{x}-b\left(\mathbf{y}\right)\right)}$
then yields the derivatives with respect to $\mathbf{w}\left(\mathbf{y}\right)$
and $b\left(\mathbf{y}\right)$ as \begin{eqnarray}
\frac{\partial D_{1}}{\partial\left(\begin{array}{c}
b\left(\mathbf{y}\right)\\
\mathbf{w}\left(\mathbf{y}\right)\end{array}\right)} & = & \frac{2}{n\, M}\int d\mathbf{x}\Pr\left(\mathbf{x}\right)\, g_{1}\left(\mathbf{x},\mathbf{y}\right)\,\left(1-Q\left(\mathbf{x}|\mathbf{y}\right)\right)\left(\begin{array}{c}
1\\
\mathbf{x}\end{array}\right)\nonumber \\
\frac{\partial D_{2}}{\partial\left(\begin{array}{c}
b\left(\mathbf{y}\right)\\
\mathbf{w}\left(\mathbf{y}\right)\end{array}\right)} & = & \frac{4\left(n-1\right)}{n\, M^{2}}\int d\mathbf{x}\Pr\left(\mathbf{x}\right)\, g_{2}\left(\mathbf{x},\mathbf{y}\right)\,\,\left(1-Q\left(\mathbf{x}|\mathbf{y}\right)\right)\left(\begin{array}{c}
1\\
\mathbf{x}\end{array}\right)\label{Eq:DerivProb}\end{eqnarray}
Because all of the properties of node $\mathbf{y}$ that are vectors
in input space (i.e. $\mathbf{x}^{\prime}\left(\mathbf{y}\right)$
and $\mathbf{w}\left(\mathbf{y}\right)$)\ are assumed to be restricted
to an input window centred on node $\mathbf{y}$, the eventual result
of evaluating the right hand sides of the above equations must be
similarly restricted to the same input window.

\subsection{The Effect of the Euclidean Norm on Minimising $D_{1}+D_{2}$}

\label{SubSect:InterpretD1D2}

The expressions for $D_{1}$ and $D_{2}$, and especially their derivatives,
are fairly complicated, so an intuitive interpretation will now be
presented. When $D_{1}+D_{2}$ is stationary with respect to variations
of $\mathbf{x}^{\prime}\left(\mathbf{y}\right)$ it may be written
as (see appendix \ref{App:D1D2RefVect}). \begin{equation}
\begin{array}{lll}
D_{1}+D_{2} & = & -\frac{2}{n}\int d\mathbf{x}\Pr\left(\mathbf{x}\right)\sum_{\mathbf{y}=\mathbf{1}}^{\mathbf{m}}\Pr\left(\mathbf{y}|\mathbf{x}\right)\left\Vert \mathbf{x}^{\prime}\left(\mathbf{y}\right)\right\Vert ^{2}\\
\\ &  & -\frac{2\left(n-1\right)}{n}\int d\mathbf{x}\Pr\left(\mathbf{x}\right)\left\Vert \sum_{\mathbf{y}=\mathbf{1}}^{\mathbf{m}}\Pr\left(\mathbf{y}|\mathbf{x}\right)\mathbf{x}^{\prime}\left(\mathbf{y}\right)\right\Vert ^{2}\\
\\ &  & +\text{ constant}\end{array}\label{Eq:D1D2RefVect}\end{equation}
The $M$ and $M^{2}$ factors do not appear in this expression because
$\Pr\left(\mathbf{y}|\mathbf{x}\right)$ is normalised to sum to unity.
The first term (which derives from $D_{1}$) is an incoherent sum
(i.e. a sum of Euclidean distances), whereas the second term (which
derives from $D_{2}$) is a coherent sum (i.e. a sum of vectors).
The first term contributes for all values of $n$, whereas the second
term contributes only for $n\geq2$, and dominates for $n\gg1$. In
order to minimise the first term the $\left\Vert \mathbf{x}^{\prime}\left(\mathbf{y}\right)\right\Vert ^{2}$
like to be as large as possible for those nodes that have a large
$\Pr\left(\mathbf{y}|\mathbf{x}\right)$. Since $\mathbf{x}^{\prime}\left(\mathbf{y}\right)$
is the centroid of the probability density $\Pr\left(\mathbf{x}|\mathbf{y}\right)$,
this implies that node $\mathbf{y}$ prefers to encode a region of
input space that is as far as possible from the origin. This is a
consequence of using a Euclidean distortion measure $\left\Vert \mathbf{x-x}^{\prime}\right\Vert ^{2}$,
which has the dimensions of $\left\Vert \mathbf{x}\right\Vert ^{2}$,
in the original definition of the distortion in equation \ref{Eq:Objective}.
In order to minimise the second term the superposition of $\mathbf{x}^{\prime}\left(\mathbf{y}\right)$
weighted by the $\Pr\left(\mathbf{y}|\mathbf{x}\right)$\ likes to
have as large a Euclidean norm as possible. Thus the nodes co-operate
amongst themselves to ensure that the nodes that have a large $\Pr\left(\mathbf{y}|\mathbf{x}\right)$
also have a large $\left\Vert \sum_{\mathbf{y}=\mathbf{1}}^{\mathbf{m}}\Pr\left(\mathbf{y}|\mathbf{x}\right)\mathbf{x}^{\prime}\left(\mathbf{y}\right)\right\Vert ^{2}$.

\section{Solvable Analytic Model}

\label{Sect:AnalyticModel}

The purpose of this section is to work through a case study in order
to demonstrate the various properties that emerge when $D_{1}+D_{2}$
is minimised.

\subsection{The Model}

\label{SubSect:Model}

It convenient to begin by ignoring the effects of leakage $\Pr\left(\mathbf{y|y}^{\prime}\right)$,
and to concentrate on a simple (non-scaling) version of the posterior
probability model (as in equation \ref{Eq:PostProb}) $\Pr\left(\mathbf{y}|\mathbf{x}\right)=\frac{Q\left(\mathbf{x}|\mathbf{y}\right)}{\sum_{\mathbf{y}^{\prime}=1}^{\mathbf{m}}Q\left(\mathbf{x}|\mathbf{y}^{\prime}\right)}$,
where the $Q\left(\mathbf{x|y}\right)$ are threshold functions of
$\mathbf{x}$\begin{equation}
Q\left(\mathbf{x|y}\right)=\left\{ \begin{array}{ll}
0 & \text{below threshold}\\
1 & \text{above threshold}\end{array}\right.\label{Eq:ThresholdFunction}\end{equation}
It is also convenient to imagine that a hypothetical infinite-sized
training set is available, so it may be described by a probability
density $\Pr\left(\mathbf{x}\right)$. This is a {}``frequentist'',
rather than a {}``Bayesian'', use of the $\Pr\left(\mathbf{x}\right)$
notation, but the distinction is not important in the context of this
paper. Assume that $\mathbf{x}=\left(\mathbf{x}_{1},\mathbf{x}_{2}\right)$
is drawn from a training set, that has 2 statistically independent
subspaces, so that \begin{equation}
\Pr\left(\mathbf{x}_{1},\mathbf{x}_{2}\right)=\Pr\left(\mathbf{x}_{1}\right)\Pr\left(\mathbf{x}_{2}\right)\label{Eq:IndependentPDF}\end{equation}
Furthermore, assume that $\Pr\left(\mathbf{x}_{1}\right)$ and $\Pr\left(\mathbf{x}_{2}\right)$
each have the form \begin{equation}
\Pr\left(\mathbf{x}_{i}\right)=\frac{1}{2\pi}\int_{0}^{2\pi}d\theta_{i}\,\delta\left(\mathbf{x}_{i}-\mathbf{x}_{i}\left(\theta_{i}\right)\right)\label{Eq:ParametricPDF}\end{equation}
i.e. $\Pr\left(\mathbf{x}_{i}\right)$ is a loop (parameterised by
a phase angle $\theta_{i}$)\ of probability density that sits in
$\mathbf{x}_{i}$-space. In order to make it easy to deduce the optimum
reference vectors, choose $\mathbf{x}_{i}\left(\theta_{i}\right)$
so that the following 2 conditions are satisfied for $i=1,2$\begin{eqnarray}
\left\Vert \mathbf{x}_{i}\left(\theta_{i}\right)\right\Vert ^{2} & = & \text{constant}\nonumber \\
\left\Vert \frac{\partial\mathbf{x}_{i}\left(\theta_{i}\right)}{\partial\theta_{i}}\right\Vert ^{2} & = & \text{constant}\label{Eq:ParametricPDFConstraint}\end{eqnarray}
This type of training set can be visualised topologically. Each training
vector $\left(\mathbf{x}_{1},\mathbf{x}_{2}\right)$ consists of 2
subvectors, each of which is parameterised by a phase angle, and which
therefore lives in a subspace that has the topology of a circle, which
is denoted as $S^{1}$. Because of the independence assumption in
equation \ref{Eq:IndependentPDF}, the pair $\left(\mathbf{x}_{1},\mathbf{x}_{2}\right)$
lives on the surface of a 2-torus, which is denoted as $S^{1}\times S^{1}$.
The minimisation of $D_{1}+D_{2}$ thus reduces to finding the optimum
way of designing an encoder/decoder for input vectors that live on
a 2-torus, with the proviso that their probability density is uniform
(this follows from equation \ref{Eq:ParametricPDF} and equation \ref{Eq:ParametricPDFConstraint}).
In order to derive the reference vectors $\mathbf{x}^{\prime}\left(\mathbf{y}\right)$,
the solution(s) of the stationarity condition $\frac{\partial\left(D_{1}+D_{2}\right)}{\partial\mathbf{x}^{\prime}\left(\mathbf{y}\right)}=0$
must be computed. The stationarity condition reduces to (see appendix
\ref{App:D1D2RefVect}) \begin{equation}
\begin{array}{l}
n\int d\mathbf{x}_{1}\, d\mathbf{x}_{2}\Pr\left(\mathbf{x}_{1},\mathbf{x}_{2}\mathbf{|y}\right)\,\left(\mathbf{x}_{1},\mathbf{x}_{2}\right)\\
\\=\left(n-1\right)\sum_{\mathbf{y}^{\prime}=\mathbf{1}}^{\mathbf{m}}\left(\int d\mathbf{x}_{1}\, d\mathbf{x}_{2}\Pr\left(\mathbf{x}_{1},\mathbf{x}_{2}|\mathbf{y}\right)\Pr\left(\mathbf{y}^{\prime}|\mathbf{x}_{1},\mathbf{x}_{2}\right)\right)\left(\mathbf{x}_{1}^{\prime}\left(y^{\prime}\right),\mathbf{x}_{2}^{\prime}\left(y^{\prime}\right)\right)\\
\\+\left(\mathbf{x}_{1}^{\prime}\left(y\right),\mathbf{x}_{2}^{\prime}\left(y\right)\right)\end{array}\label{Eq:StationaryRefVect}\end{equation}

\begin{figure}
\centering{}\label{Fig:S1S1}\includegraphics[clip,width=10cm]{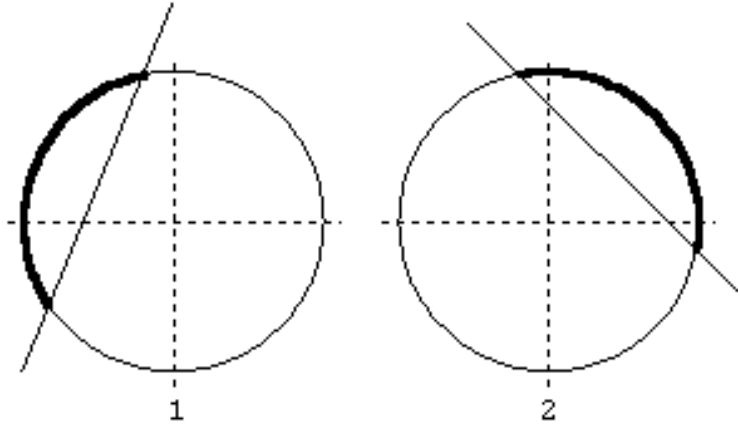}\caption{Representation of $S^{1}\times S^{1}$ topology with a threshold $Q\left(x|y\right)$
superimposed.}

\end{figure}

\begin{figure}
\begin{centering}
\label{Fig:Torus}\includegraphics[clip,width=10cm]{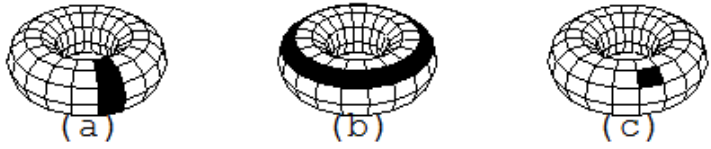}
\par\end{centering}

\centering{}\caption{Explicit representation of $S^{1}\times S^{1}$ topology as a torus
with the effect of 3 different types of threshold $Q\left(x|y\right)$
shown.}

\end{figure}

It is useful to use the simple diagrammatic notation shown in figure
\ref{Fig:S1S1}. %
Each circle in figure \ref{Fig:S1S1} represents one of the $S^{1}$
subspaces, so the two circles together represent the product $S^{1}\times S^{1}$.
The constraints in equation \ref{Eq:ParametricPDFConstraint} are
represented by each circle being centred on the origin of its subspace
($\left\Vert \mathbf{x}_{i}\left(\theta_{i}\right)\right\Vert ^{2}$
is constant), and the probability density around each circle being
constant ($\left\Vert \frac{\partial\mathbf{x}_{i}\left(\theta_{i}\right)}{\partial\theta_{i}}\right\Vert ^{2}$
is constant). A single threshold function $Q\left(\mathbf{x|y}\right)$
is represented by a chord cutting through each circle (with 0 and
1 indicating on which side of the chord the threshold is triggered).
The $\mathbf{x}_{i}$ that lie above threshold in each subspace are
highlighted. Both $\mathbf{x}_{1}$ and $\mathbf{x}_{2}$ must lie
above threshold in order to ensure $Q\left(\mathbf{x|y}\right)=1$,
i.e. they must both lie within regions that are highlighted in figure
\ref{Fig:S1S1}. In this case node $\mathbf{y}$ will be said to be
{}``attached'' to both subspace 1 and subspace 2. A special case
arises when the chord in one of the subspaces (say it is $\mathbf{x}_{2}$)\
does not intersect the circle at all, and the circle lies on the side
of the chord where the threshold is triggered. In this case $Q\left(\mathbf{x|y}\right)$
does not depend on $\mathbf{x}_{2}$, so that $\Pr\left(\mathbf{y|x}_{1},\mathbf{x}_{2}\right)=\Pr\left(\mathbf{y|x}_{1}\right)$,
in which case node $\mathbf{y}$ will be said to be {}``attached''
to subspace 1 but {}``detached'' from subspace 2. The typical ways
in which a node becomes attached to the 2-torus are shown in figure
\ref{Fig:Torus}. %
In figure \ref{Fig:Torus}(a) the node is attached to one of the $S^{1}$
subspaces and detached from the other. In figure \ref{Fig:Torus}(b)
the attached and detached subspaces are interchanged with respect
to figure \ref{Fig:Torus}(a). In figure \ref{Fig:Torus}(c) the node
is attached to both subspaces.

\subsection{All Nodes Attached to One Subspace}

\label{SubSect:AttachOne}

\begin{figure}
\begin{centering}
\label{Fig:AttachOne}\includegraphics[clip,width=10cm]{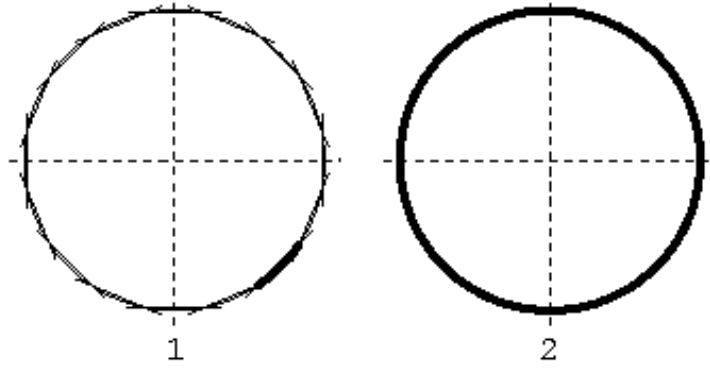}
\par\end{centering}

\caption{16 nodes are shown, which are all attached to subspace 1, and all
detached from subspace 2.}

\end{figure}

Consider the configuration of threshold functions shown in figure
\ref{Fig:AttachOne}. This is equivalent to all of the nodes being
attached to loops to cover the 2-torus, with a typical node being
as shown in figure \ref{Fig:Torus}(a) (or, equivalently, figure \ref{Fig:Torus}(b)).
When $D_{1}+D_{2}$ is minimised, it is assumed that the 4 nodes are
symmetrically disposed in subspace 1, as shown. Each is triggered
if and only if $\mathbf{x}_{1}$ lies within its quadrant, and one
such quadrant is highlighted in figure \ref{Fig:AttachOne}. This
implies that only 1 node is triggered at a time. The assumed form
of the threshold functions implies $\Pr\left(y\mathbf{|x}_{1},\mathbf{x}_{2}\right)=\Pr\left(y\mathbf{|x}_{1}\right)$,
so equation \ref{Eq:StationaryRefVect} reduces to \begin{equation}
\begin{array}{l}
n\int d\mathbf{x}_{1}\, d\mathbf{x}_{2}\Pr\left(\mathbf{x}_{1}|y\right)\Pr\left(\mathbf{x}_{2}\right)\,\left(\mathbf{x}_{1},\mathbf{x}_{2}\right)\\
\\=\int d\mathbf{x}_{1}\, d\mathbf{x}_{2}\Pr\left(\mathbf{x}_{1}|y\right)\Pr\left(\mathbf{x}_{2}\right)\\
\\\times\left(\left(n-1\right)\sum_{y^{\prime}=1}^{M}\Pr\left(y^{\prime}\mathbf{|x}_{1}\right)\left(\mathbf{x}_{1}^{\prime}\left(y^{\prime}\right),\mathbf{x}_{2}^{\prime}\left(y^{\prime}\right)\right)+\left(\mathbf{x}_{1}^{\prime}\left(y\right),\mathbf{x}_{2}^{\prime}\left(y\right)\right)\right)\end{array}\ \ \ \ \label{Eq:StationaryOne}\end{equation}
 whence \begin{eqnarray}
\mathbf{x}_{1}^{\prime}\left(y\right) & = & \int d\mathbf{x}_{1}\Pr\left(\mathbf{x}_{1}|y\right)\,\mathbf{x}_{1}\nonumber \\
\mathbf{x}_{2}^{\prime}\left(y\right) & = & 0\label{Eq:RefVectOne}\end{eqnarray}

\subsection{All Nodes Attached to Both Subspaces}

\label{SubSect:AttachBoth}

\begin{figure}
\begin{centering}
\label{Fig:AttachBoth}\includegraphics[clip,width=10cm]{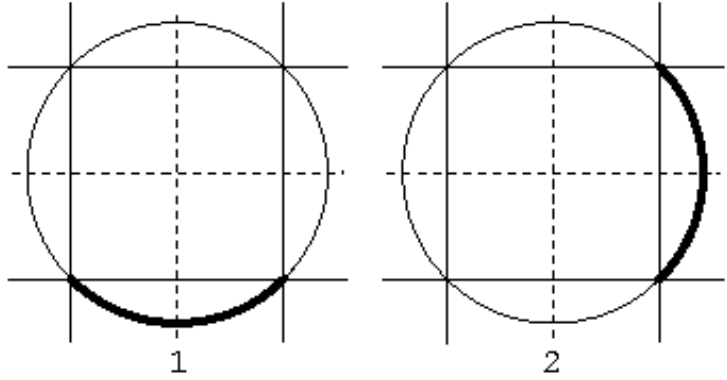}
\par\end{centering}

\caption{16 nodes are shown, which are all attached to both subspace 1 and
subspace 2.}

\end{figure}

Consider the configuration of threshold functions shown in figure
\ref{Fig:AttachBoth}. This is equivalent to all of the nodes being
attached to patches to cover the 2-torus, with a typical node being
as shown in figure \ref{Fig:Torus}(c). %
In this case, when $D_{1}+D_{2}$ is minimised, it is assumed that
each subspace is split into 2 halves. This requires a total of 4 nodes,
each of which is triggered if, and only if, both $\mathbf{x}_{1}$
and $\mathbf{x}_{2}$ lie on the corresponding half-circles. This
implies that only 1 node is triggered at a time. The assumed form
of the threshold functions implies that the stationarity condition
becomes \begin{equation}
\begin{array}{l}
n\,\Pr\left(y\right)\int d\mathbf{x}_{1}\, d\mathbf{x}_{2}\Pr\left(\mathbf{x}_{1},\mathbf{x}_{2}|y\right)\,\left(\mathbf{x}_{1},\mathbf{x}_{2}\right)\\
\\=\Pr\left(y\right)\int d\mathbf{x}_{1}\, d\mathbf{x}_{2}\Pr\left(\mathbf{x}_{1},\mathbf{x}_{2}|y\right)\\
\\\times\left(\left(n-1\right)\sum_{y^{\prime}=1}^{M}\Pr\left(y^{\prime}\mathbf{|x}_{1},\mathbf{x}_{2}\right)\left(\mathbf{x}_{1}^{\prime}\left(y^{\prime}\right),\mathbf{x}_{2}^{\prime}\left(y^{\prime}\right)\right)+\left(\mathbf{x}_{1}^{\prime}\left(y\right),\mathbf{x}_{2}^{\prime}\left(y\right)\right)\right)\end{array}\ \ \ \ \ \end{equation}
whence \begin{eqnarray}
\mathbf{x}_{1}^{\prime}\left(y\right) & = & \int d\mathbf{x}_{1}\Pr\left(\mathbf{x}_{1}|y\right)\mathbf{x}_{1}\nonumber \\
\mathbf{x}_{2}^{\prime}\left(y\right) & = & \int d\mathbf{x}_{2}\Pr\left(\mathbf{x}_{2}|y\right)\mathbf{x}_{2}\label{Eq:RefVectBoth}\end{eqnarray}

\subsection{Half the Nodes Attached to One Subspace, and Half to the Other Subspace}

\label{SubSect:AttachEither}

\begin{figure}
\begin{centering}
\label{Fig:AttachEither}\includegraphics[clip,width=10cm]{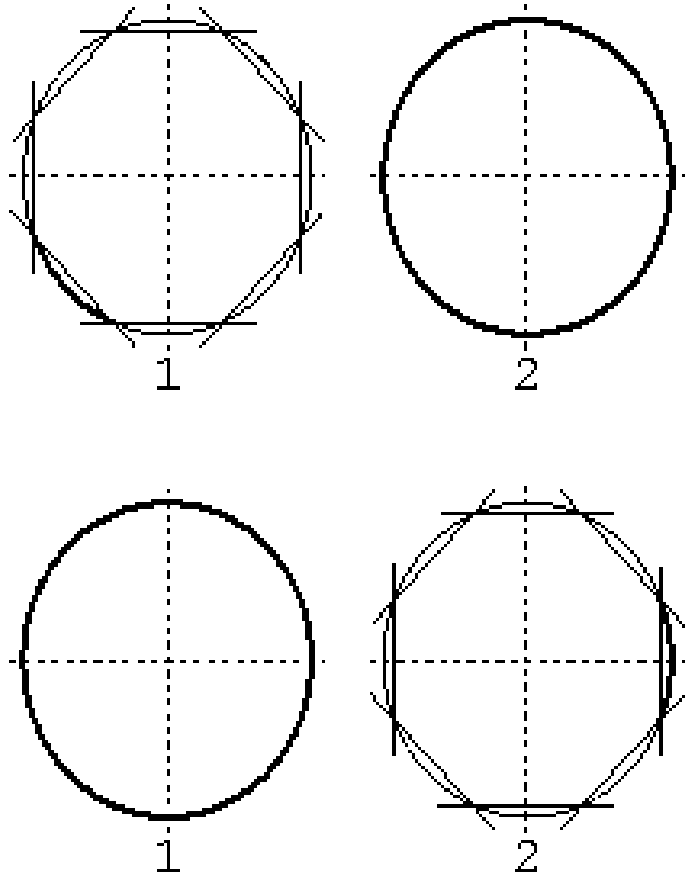}
\par\end{centering}

\caption{16 nodes are shown, 8 of which are attached to subspace 1 and detached
from subspace 2 (top row), and 8 of which are attached to subspace
2 and detached from subspace 1 (bottom row).}

\end{figure}

Consider the configuration of threshold functions shown in figure
\ref{Fig:AttachEither}. This is equivalent to half of the nodes being
attached to loops to cover the 2-torus, with a typical node being
as shown in figure \ref{Fig:Torus}(a). The other half of the nodes
would then be attached in an analogous way, but as shown in figure
\ref{Fig:Torus}(b). Thus the 2-torus is covered twice over. %
In this case, when $D_{1}+D_{2}$ is minimised, it is assumed that
each subspace is split into 2 halves. This requires a total of 4 nodes,
each of which is triggered if $\mathbf{x}_{1}$ (or $\mathbf{x}_{2}$)
lies on the half-circle in the subspace to which the node is attached.
Thus exactly 2 nodes $y_{1}\left(\mathbf{x}_{1}\right)$ and $y_{2}\left(\mathbf{x}_{2}\right)$
are triggered at a time, so that \begin{eqnarray}
\Pr\left(y|\mathbf{x}_{1},\mathbf{x}_{2}\right) & = & \frac{1}{2}\left(\delta_{y,y_{1}\left(\mathbf{x}_{1}\right)}+\delta_{y,y_{2}\left(\mathbf{x}_{2}\right)}\right)\nonumber \\
\  & = & \frac{1}{2}\left(\Pr\left(y|\mathbf{x}_{1}\right)+\Pr\left(y|\mathbf{x}_{2}\right)\right)\end{eqnarray}
For simplicity, assume that node $y$ is attached to subspace 1, then
$\Pr\left(\mathbf{x}_{1},\mathbf{x}_{2}|y\right)=\Pr\left(\mathbf{x}_{1}|y\right)\Pr\left(\mathbf{x}_{2}\right)$
and the stationarity condition becomes \begin{equation}
\begin{array}{l}
n\,\Pr\left(y\right)\int d\mathbf{x}_{1}\, d\mathbf{x}_{2}\Pr\left(\mathbf{x}_{1}|y\right)\Pr\left(\mathbf{x}_{2}\right)\,\left(\mathbf{x}_{1},\mathbf{x}_{2}\right)\\
\\=\Pr\left(y\right)\int d\mathbf{x}_{1}\, d\mathbf{x}_{2}\Pr\left(\mathbf{x}_{1}|y\right)\Pr\left(\mathbf{x}_{2}\right)\\
\\\times\left(\begin{array}{c}
\frac{n-1}{2}\sum_{y^{\prime}=1}^{M}\left(\Pr\left(y^{\prime}|\mathbf{x}_{1}\right)+\Pr\left(y^{\prime}|\mathbf{x}_{2}\right)\right)\,\left(\mathbf{x}_{1}^{\prime}\left(y^{\prime}\right),\mathbf{x}_{2}^{\prime}\left(y^{\prime}\right)\right)\\
+\left(\mathbf{x}_{1}^{\prime}\left(y\right),\mathbf{x}_{2}^{\prime}\left(y\right)\right)\end{array}\right)\end{array}\ \ \ \ \ \ \end{equation}
This may be simplified to yield \begin{equation}
\begin{array}{l}
\ n\int d\mathbf{x}_{1}\Pr\left(\mathbf{x}_{1}|y\right)\,\left(\mathbf{x}_{1},0\right)\\
\\=\frac{n+1}{2}\left(\mathbf{x}_{1}^{\prime}\left(y\right),\mathbf{x}_{2}^{\prime}\left(y\right)\right)\\
\\+\frac{n-1}{2}\int d\mathbf{x}_{2}\Pr\left(\mathbf{x}_{2}\right)\sum_{y^{\prime}=1}^{M}\Pr\left(y^{\prime}|\mathbf{x}_{2}\right)\,\left(\mathbf{x}_{1}^{\prime}\left(y^{\prime}\right),\mathbf{x}_{2}^{\prime}\left(y^{\prime}\right)\right)\\
\\=\frac{n+1}{2}\left(\mathbf{x}_{1}^{\prime}\left(y\right),\mathbf{x}_{2}^{\prime}\left(y\right)\right)+\frac{n-1}{2}\left\langle \mathbf{x}_{1}^{\prime}\left(y\right),\mathbf{x}_{2}^{\prime}\left(y\right)\right\rangle _{2}\end{array}\ \end{equation}
Write the 2 subspaces separately (remember that node $y$ is assumed
to be attached to subspace 1) \begin{eqnarray}
\mathbf{x}_{1}^{\prime}\left(y\right) & = & \frac{2n}{n+1}\int d\mathbf{x}_{1}\Pr\left(\mathbf{x}_{1}|y\right)\mathbf{x}_{1}-\frac{n-1}{n+1}\left\langle \mathbf{x}_{1}^{\prime}\left(y\right)\right\rangle _{2}\nonumber \\
\mathbf{x}_{2}^{\prime}\left(y\right) & = & -\frac{n-1}{n+1}\left\langle \mathbf{x}_{2}^{\prime}\left(y\right)\right\rangle _{2}\end{eqnarray}
If this result is simultaneously solved with the analogous result
for node $y$ attached to subspace 2, then the $\left\langle \cdots\right\rangle $
terms vanish to yield \begin{eqnarray}
\mathbf{x}_{1}^{\prime}\left(y\right) & = & \left\{ \begin{array}{ll}
\frac{2n}{n+1}\int d\mathbf{x}_{1}\Pr\left(\mathbf{x}_{1}|y\right)\mathbf{x}_{1} & y\text{ attached to subspace 1}\\
0 & y\text{ attached to subspace 2}\end{array}\right.\nonumber \\
\mathbf{x}_{2}^{\prime}\left(y\right) & = & \left\{ \begin{array}{ll}
0 & y\text{ attached to subspace 1}\\
\frac{2n}{n+1}\int d\mathbf{x}_{2}\Pr\left(\mathbf{x}_{2}|y\right)\mathbf{x}_{2} & y\text{ attached to subspace 2}\end{array}\right.\label{Eq:RefVectEither}\end{eqnarray}

\subsection{Compare $D_{1}+D_{2}$ for the 3 Different Types of Solution}

\label{SubSect:CompareD1D2}

\begin{figure}
\begin{centering}
\label{Fig:PlotN1}\includegraphics[bb=0bp 0bp 1067bp 749bp,width=10cm]{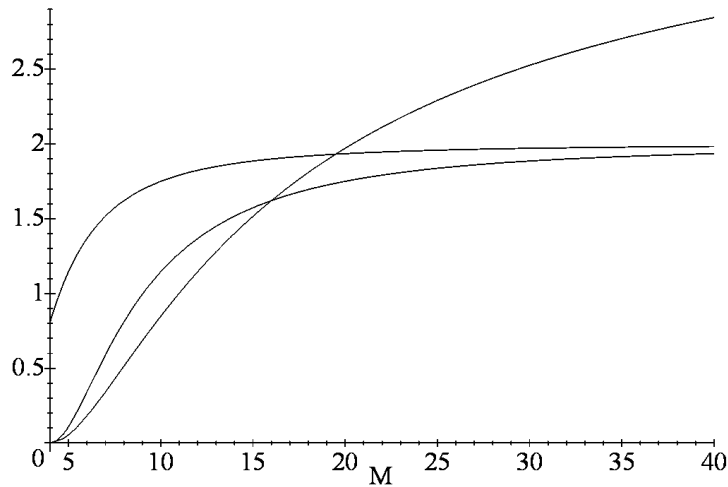}
\par\end{centering}

\caption{Plots of $-D_{1}-D_{2}$ for $n=1$ for each of the 3 types of optimum.}

\end{figure}

Consider the left hand side of figure \ref{Fig:AttachOne} for the
case of $M$ nodes, when the $M$ threshold functions form a regular
$M$-ogon. $\Pr\left(\mathbf{x|}y\right)$ then denotes the part of
the circle that is associated with node $y$, whose radius of gyration
squared is given by (assuming that the circle has unit radius) \begin{eqnarray}
R_{M} & \equiv & \left\Vert \int d\mathbf{x}\Pr\left(\mathbf{x|}y\right)\,\mathbf{x}\right\Vert ^{2}\nonumber \\
 & = & \left(\frac{M}{2\pi}\int_{0}^{\frac{2\pi}{M}}d\theta\cos\theta\right)^{2}\nonumber \\
 & = & \left(\frac{M}{2\pi}\sin\frac{2\pi}{M}\right)^{2}\label{Eq:RadiusSquared}\end{eqnarray}
Gather the results for $\left(\mathbf{x}_{1}^{\prime}\left(y\right),\mathbf{x}_{2}^{\prime}\left(y\right)\right)$
in equations \ref{Eq:RefVectOne} (referred to as type 1), \ref{Eq:RefVectBoth}
(referred to as type 2), and \ref{Eq:RefVectEither} (referred to
as type 3) together and insert them into $D_{1}+D_{2}$ in equation
\ref{Eq:D1D2RefVect} to obtain (see appendix \ref{App:D1D2Compare})

\begin{equation}
D_{1}+D_{2}=\left\{ \begin{array}{ll}
\text{constant}-2R_{M} & \text{type 1}\\
\text{constant}-4R_{\sqrt{M}} & \text{type 2}\\
\text{constant}-\frac{4n}{n+1}R_{\frac{M}{2}} & \text{type 3}\end{array}\right.\end{equation}
 In figure \ref{Fig:PlotN1} the 3 solutions are plotted for the case
$n=1$. %
For $n=1$ the type 3 solution is never optimal, the type 1 solution
is optimal for $M\leq19$, and the type 2 solution is optimal for
$M\geq20.$ This behaviour is intuitively sensible, because a larger
number of nodes is required to cover a 2-torus as shown in figure
\ref{Fig:Torus}(c) than as shown in figure \ref{Fig:Torus}(a) (or
figure \ref{Fig:Torus}(b)).

\begin{figure}
\begin{centering}
\label{Fig:PlotN2}\includegraphics[bb=0bp 0bp 1075bp 743bp,width=10cm]{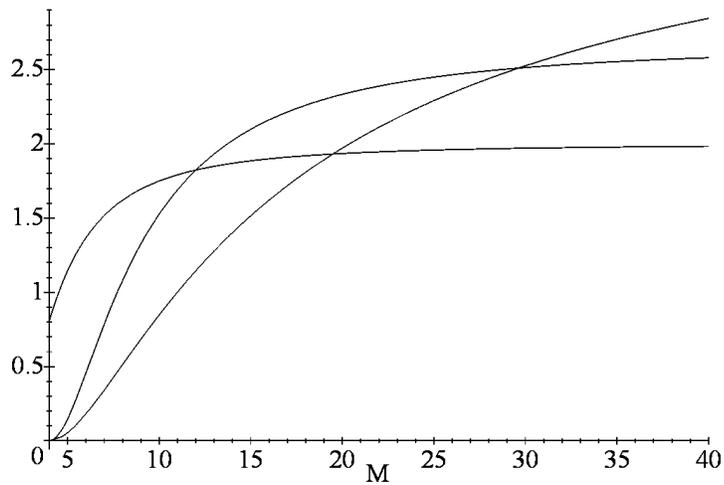}
\par\end{centering}

\caption{Plots of $-D_{1}-D_{2}$ for $n=2$ for each of the 3 types of optimum.}

\end{figure}

In figure \ref{Fig:PlotN2} the 3 solutions are plotted for the case
$n=2$. %
For $n=2$ the type 1 solution is optimal for $M\leq12$, and the
type 2 solution is optimal for large $M\geq30$, but there is now
an intermediate region $12\leq M\leq29$ (type 1 and type 3 have an
equal $D_{1}+D_{2}$ at $M=12$) where the $n$-dependence of the
type 3 solution has now made it optimal. Again, this behaviour is
intuitively reasonable, because the type 3 solution requires at least
2 observations in order to be able to yield a small Euclidean resonstruction
error in each of the 2 subspaces, i.e. for $n=2$ the 2 nodes that
fire must be attached to different subspaces. Note that in the type
3 solution the nodes that fire are not guaranteed to be attached to
different subspaces. In the type 3 solution there is a probability
$\frac{1}{2^{n}}\frac{n!}{n_{1}!n_{2}!}$ that $n_{i}$ (where $n=n_{1}+n_{2}$)
nodes are attached to subspace $i$, so the trend is for the type
3 solution to become more favoured as $n$ is increased.

\begin{figure}
\begin{centering}
\label{Fig:PlotNAsymptotic}\includegraphics[bb=0bp 0bp 1057bp 741bp,width=10cm]{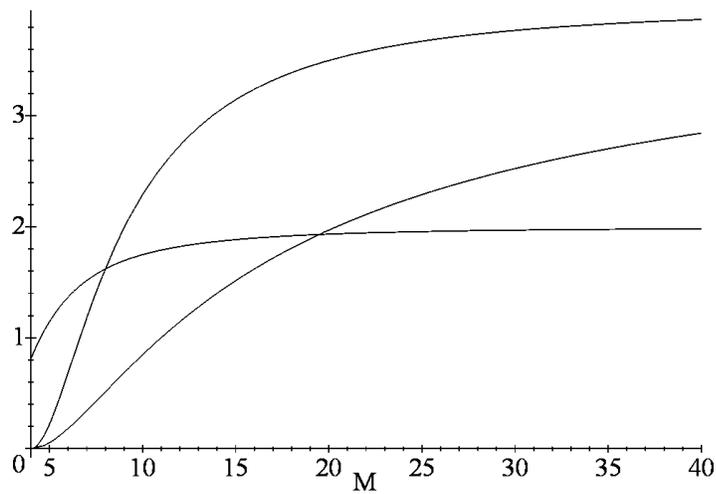}
\par\end{centering}

\caption{Plots of $-D_{1}-D_{2}$ for $n\rightarrow\infty$ for each of the
3 types of optimum.}

\end{figure}

\begin{figure}

\centering{}\label{Fig:PhaseDiagram}\includegraphics[clip,width=10cm]{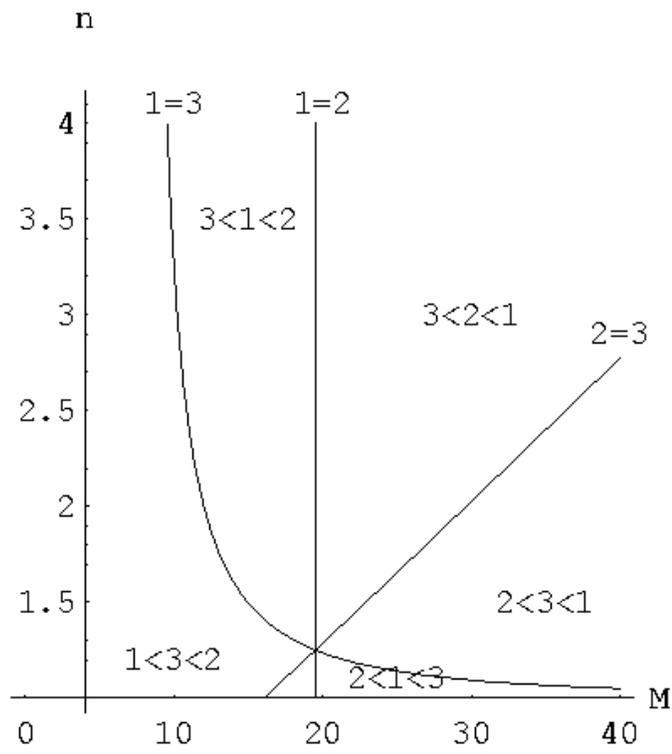}\caption{Phase diagram of the regions in which the 3 types of solution are
optimal.}

\end{figure}

In figure \ref{Fig:PlotNAsymptotic} the 3 solutions are plotted for
the case $n\rightarrow\infty$. %
For $n\rightarrow\infty\,$ the type 2 solution is never optimal ,
the type 1 solution is optimal for $M\leq8$, and the type 3 solution
is optimal for $M\geq8$. The type 2 solution approaches the type
3 solution from below asymptotically as $M\rightarrow\infty$. In
figure \ref{Fig:PhaseDiagram} a phase diagram is given which shows
how the relative stability of the 3 types of solution for different
$M$ and $n$, where the type 3 solution is seen to be optimal over
a large part of the $\left(M,n\right)$ plane. %
Thus the most interesting, and commonly occurring, solution is the
one in which half the nodes are attached to one subspace and half
to the other subspace (i.e. solution type 3). Although this result
has been derived using the non-scaling version of the posterior probability
model $\Pr\left(\mathbf{y|x}\right)$ (as in equation \ref{Eq:PostProb}),
it may also be used for scaling posterior probabilities (as given
in equation \ref{Eq:PostProbPMD}) in certain limiting cases, and
also for cases where the effect of leakage $\Pr\left(\mathbf{y|y}^{\prime}\right)$
is small.

\subsection{Various Extensions}

\subsubsection{The Effect of Leakage}

The effect of leakage will not be analysed in detail here. However,
its effect may readily be discussed phenomenologically, because the
optimisation acts to minimise the damaging effect of leakage on the
posterior probability by ensuring that the properties of nodes that
are connected by leakage are similar. This has the most dramatic effect
on the type 3 solution, where the way in which the nodes are partitioned
into 2 halves must be very carefully chosen in order to minimise the
damage due to leakage. If the leakage is presumed to be a local function,
so that $\Pr\left(y|y^{\prime}\right)=\pi\left(y-y^{\prime}\right)$,
which is a localised {}``blob''-shaped function, then the properties
of adjacent node are similar (after optimisation). Since nodes that
are attached to 2 different subspaces necessarily have very different
properties, whereas nodes that are attached to the same subspace can
have similar properties, it follows that the nodes must split into
2 continguous halves, where nodes $1,2,\cdots,\frac{M}{2}$ are attached
to subspace 1 and nodes $\frac{M}{2}+1,\frac{M}{2}+2,\cdots,M$ are
attached to subspace 2, or vice versa. The effect of leakage is thereby
minimised, with the worst effect occurring at the boundary between
the 2 halves of nodes.

\subsubsection{Modifying the Posterior Probability to Become Scalable}

The above analysis has focussed on the non-scaling version of the
posterior probability, in which all $M$ nodes act together as a unit.
The more general scaling case where the $M$ nodes are split up by
the effect of the neighbourhood function $N\left(y\right)$ will not
be analysed in detail, because many of its properties are essentially
the same as in the non-scaling case. For simplicity assume that the
neighbourhood function $N\left(y\right)$ is a {}``top-hat'' with
width $w$ (an odd integer) centred on $y$. Impose periodic boundary
conditions so that the inverse neighbourhood function $\tilde{N}\left(y\right)$
is also a top-hat, $\tilde{N}\left(y\right)=N\left(y\right)$. In
this case an optimum solution in the non-scaling case (with $M=w$)\ can
be directly related to a corresponding optimum solution in the scaling
case by simply repeating the node properties periodically every $w$
nodes. Strictly speaking, higher order periodicities can also occur
in the scaling case (and can be favoured under certain conditions),
where the period is $\frac{w}{k}$ ($k$ is an integer), but these
will not be discussed here.

The effect of the periodic replication of node properties is interesting.
The type 3 solution (with leakage and with $M=w)$ splits the nodes
into 2 halves, where nodes $1,2,\cdots,\frac{w}{2}$ are attached
to subspace 1 and nodes $\frac{w}{2}+1,\frac{w}{2}+2,\cdots,w$ are
attached to subspace 2, or vice versa. When this is replicated periodically
every $w$ nodes it produces an alternating structure of node properties,
where $\frac{w}{2}$ nodes are attached to subspace 1, then the next
$\frac{w}{2}$ nodes are attached to subspace 2, and thenthe next
$\frac{w}{2}$ nodes are attached to subspace 1, and so on. This behaviour
is reminiscent of the so-called {}``dominance stripes'' that are
observed in the mammalian visual cortex.

\section{Experiment}

\label{Sect:Experiment}

The purpose of this section is to demonstrate the emergence of the
dominance stripes in numerical simulations. The main body of the software
is concerned with evaluating the derivatives of $D_{1}+D_{2}$, and
the main difficulty is choosing an appropriate form for the leakage
(this has not yet been automated).

\subsection{The Parameters}

The parameters that are required for a simulation are as follows:
\begin{enumerate}
\item $\left(m_{1},m_{2}\right)$: size of 2D rectangular array of nodes.
$M=m_{1}m_{2}$.
\item $\left(i_{1},i_{2}\right)$: size of 2D rectangular input window for
each node (odd integers). Ensure that the input window is not too
many input data {}``correlation areas'' in size, otherwise dominance
stripes may not emerge. Dominance stripes require that the correlation
\textit{within} an input window are substantially stronger than the
correlations \textit{between} input windows that are attached to different
subspaces.
\item $\left(w_{1},w_{2}\right)$: size of 2D rectangular neighbourhood
window for each node (odd integers). The neighbourhood function $N\left(y_{1},y_{2}\right)$
is a rectangular top-hat centred on $\left(y_{1},y_{2}\right)$. The
size of the neighbourhood window has to lie within a limited range
to ensure that dominance stripes are produced. This corresponds to
ensuring that $M$ lies in the type 3 region of the phase diagram
in figure \ref{Fig:PhaseDiagram}. It is also preferable for the size
of the neighbourhood window to be substantially smaller than the input
window, otherwise different parts of a neighbourhood window will see
different parts of the input data, which will make the network behaviour
more difficult to interpret.
\item $\left(l_{1},l_{2}\right)$: size of 2D rectangular leakage window
for each node (odd integers). For simplicity the leakage $\Pr\left(\mathbf{y|y}^{\prime}\right)$
is assumed to be given by $\Pr\left(\mathbf{y|y}^{\prime}\right)=\pi\left(\mathbf{y}-\mathbf{y}^{\prime}\right)$,
where $\pi\left(\mathbf{y}-\mathbf{y}^{\prime}\right)$ is a {}``top-hat''
function of $\mathbf{y}-\mathbf{y}^{\prime}$ which covers a rectangular
region of size $\left(l_{1},l_{2}\right)$ centred on $\mathbf{y}-\mathbf{y}^{\prime}=0$\textbf{.
}The size of the leakage window must be large enough to correlate
the parameters of adjacent nodes, but not so large that it enforces
such strong correlations between the node parameters that it destroys
dominance stripes.
\item $\nu$: additive noise level used to corrupt each member of the training
set.
\item $\kappa$: wavenumber of sinusoids used in the training set. In describing
the training sets the index $y$ will be used to denote position in
input space, thus position $y$ in input space lies directly {}``under''
node $y$ of the network. In 1D simulations each training vector is
a sinusoid of the form $\sin\left(\kappa y+\phi\right)+r$, where
$\phi$ is a random phase angle, and $r$ is a random number sampled
uniformly from the interval $\left[-\frac{\nu}{2},\frac{\nu}{2}\right]$
- this generates an $S^{1}$ topology training set (i.e. parameterised
by 1 random angle). In 2D simulations each training vector is a sinusoid
of the form $\sin\left(\kappa\left(y_{1}\cos\theta+y_{2}\sin\theta\right)+\phi\right)+r$,
where the additional angle $\theta$ is a random azimuthal orientation
for the sine wave - this generates an $S^{1}\times S^{1}$ topology
training set (i.e. parameterised by 2 independent random angles).
Note that $\kappa i_{1}$ and $\kappa i_{2}$ must be an integer multiple
of $2\pi$ in order to ensure that the probability density around
the $S^{1}$ subspace generated by $\phi$ has uniform density (in
effect, the $S^{1}$ then becomes a circular Lissajous figure, which
therefore has uniform probability density, unlike non-circular Lissajous
figures), and thus to ensure that there are no artefacts induced by
the periodicity of the training data that might mimic the effect of
dominance stripes. If $\kappa i_{1}$ and $\kappa i_{2}$ are much
greater than $2\pi$ then it is not necessary to fix them to be integer
multiples of $2\pi$ - because the fluctuations in the probability
density are then negligible. Note that this restriction on the value
of $\kappa$ would not have been necessary had complex exponentials
been used rather than sinusoids.
\item $s$: number of subspaces. This fixes the number of statistically
independent subspaces in the training set. When $s=1$ the training
set is generated exactly as above. When $s=2$ the training set is
split up as follows. The 1D case has even $y$ in one subspace, and
odd $y$ in the other subspace, thus successive components of each
training vector alternate between the 2 subspaces. The 2D case has
even $y_{1}+y_{2}$ in one subspace, and odd $y_{1}+y_{2}$ in the
other subspace, thus each training vector is split up into a chessboard
pattern of interlocking subspaces. This strategy readily generalises
for $s\geq3$, although this is not used here. Within each subspace
the training vector is generated as above, and the subspaces are generated
so that they are statistically independent.
\item $\varepsilon$: update parameter used in gradient descent. This is
used to update parameters thus \begin{equation}
\text{parameter}\rightarrow\text{parameter}-\varepsilon\,\frac{\partial\left(D_{1}+D_{2}\right)}{\partial\text{parameter}}\label{eqUpdatePrescription}\end{equation}
 There are 3 internally generated update parameter, which control
the update of the 3 different types of parameter, i.e. the biases,
the weights, and the reference vectors. This is necessary because
these parameters all have different dimensionalities, and by inspection
of equation \ref{eqUpdatePrescription} the dimensionality of an update
parameter is the dimensionality of the parameter it updates (squared)
divided by the dimensionality of the Euclidean distortion. These 3
internal parameters are automatically adjusted to ensure that the
average change in absolute value of each of the 3 types of parameter
is equal to $\varepsilon$ times the typical diameter of the region
of parameter space populated by the parameters. This adjustment is
made anew as each training vector is presented. The size of $\varepsilon$
determines the {}``memory time'' of the node parameters. This memory
time determines the effective number of training vectors that the
nodes are being optimised against, and thus must be sufficiently long
(i.e. $\varepsilon$ sufficiently small) that if $s\geq2$ it is possible
to discern that the subspaces are indeed statistically independent.
This is crucially important, for dominance stripes cannot be obtained
if the subspaces are not sufficiently statistically independent. So
$\varepsilon$ must be small, which unfortunately leads to correspondingly
long training times. 
\end{enumerate}

\subsection{Initialisation}

The training set is globally translated and scaled so that the components
of all of its training vectors lie in the interval $[-1,+1]$. There
are 3 parameter types to initialise. The weights were all initialised
to random numbers sampled from a uniform distribution in the interval
$\left[-0.1,+0.1\right]$, whereas the biasses and the reference vector
components were all initialised to 0. Because the 2D simulations took
a very long time to run, they were periodically interrupted and the
state of all the variables written to an output file. The simulation
could then be continued by reading this output file in again and simply
continuing where the simulation left off. Alternatively, some of the
variables might have their values changed before continuing. In particular,
the random number generator could thus be manipulated to simulate
the effect of a finite sized training set (i.e. use the \textit{same}
random number seed at the start of each part of the simulation), or
an infinite-sized training set (i.e. use a \textit{different} random
number seed at the start of each part of the simulation). The size
of the $\varepsilon$ parameter could also thus be manipulated should
a large value be required initially, and reduced to a small value
later on, as required in order to guarantee that when $n\geq2$ the
input subspaces are seen to be statistically independent, and dominance
stripes may emerge.

\subsection{Boundary Conditions}

There are many ways to choose the boundary conditions. In the numerical
simulations periodic boundary conditions will be avoided, because
they can lead to artefacts in which the node parameters become topologically
trapped. For instance, in a 2D simulation, periodic boundary conditions
imply that the nodes sit on a 2-torus. Leakage implies that the node
parameter values are similar for adjacent nodes, which limits the
freedom for the parameters to adjust their values on the surface of
the 2-torus. For instance, any acceptable set of parameters that sits
on the 2-torus can be converted into another acceptable set by mapping
the 2-torus to itself, so that each of its $S^{1}$ {}``coils up''
an integer number of times onto itself. Such a multiply wrapped parameter
configuration is topologically trapped, and cannot be perturbed to
its original form. This problem does not arise with non-periodic boundary
conditions.

There are several different problems that arise at the boundaries
of the array of nodes:
\begin{enumerate}
\item The neighbourhood function $N\left(y_{1},y_{2}\right)$ cannot be
assumed to be a rectangular top-hat centred on $\left(y_{1},y_{2}\right)$.
Instead, it will simply be truncated so that it does not fall off
the edge of array of nodes, i.e. $N\left(y_{1},y_{2}\right)=0$ for
those $\left(y_{1},y_{2}\right)$ that lie outside the array.
\item The leakage function $\pi\left(y_{1}-y_{1}^{\prime},y_{2}-y_{2}^{\prime}\right)$
will be similarly truncated. However, in this case $\pi\left(y_{1}-y_{1}^{\prime},y_{2}-y_{2}^{\prime}\right)$
must normalise to unity when summed over $\left(y_{1},y_{2}\right)$,
so the effect of the truncation must be compensated by scaling the
remaining elements of $\pi\left(y_{1}-y_{1}^{\prime},y_{2}-y_{2}^{\prime}\right)$.
\item The input window for each node implies that the input array must be
larger than the node array in order that the input windows never fall
off the edge of the input array. 
\end{enumerate}

\subsection{Presentation of Results}

The most important result is the emergence of dominance stripes. For
$n=2$ there are thus 2 numbers that need to be displayed for each
node: the {}``degree of attachment'' to subspace 1, and similarly
for subspace 2. There are many ways to measure degree of attachment,
for instance the probability density $\Pr\left(\mathbf{x|y}\right)$
gives a direct measurement of how strongly node $\mathbf{y}$ depends
on the input vector $\mathbf{x}$, so its {}``width'' or {}``volume''
in each of the subspaces could be used to measure degree of attachment.
However, in the simulations presented here (i.e. sinusoidal training
vectors) the degree of attachment is measured as the average of the
absolute values of the components of the reference vector in the subspace
concerned. This measure tends to zero for complete detachment. For
1D simulations 2 dominance plots can be overlaid to show the dominance
of subspaces 1 and 2 for each node. For 2D simulations it is simplest
to present only 1 of these plots as a 2D array of grey-scale pixels,
where the grey level indicates the dominance of subspace 1 (or, alternatively,
subspace 2).%
\footnote{The results of the 2D simulations do not appear in this draft paper.%
}

\subsection{1D Simulation}

\begin{figure}

\begin{centering}
\label{Fig:Dominance1D}\includegraphics[clip,width=10cm]{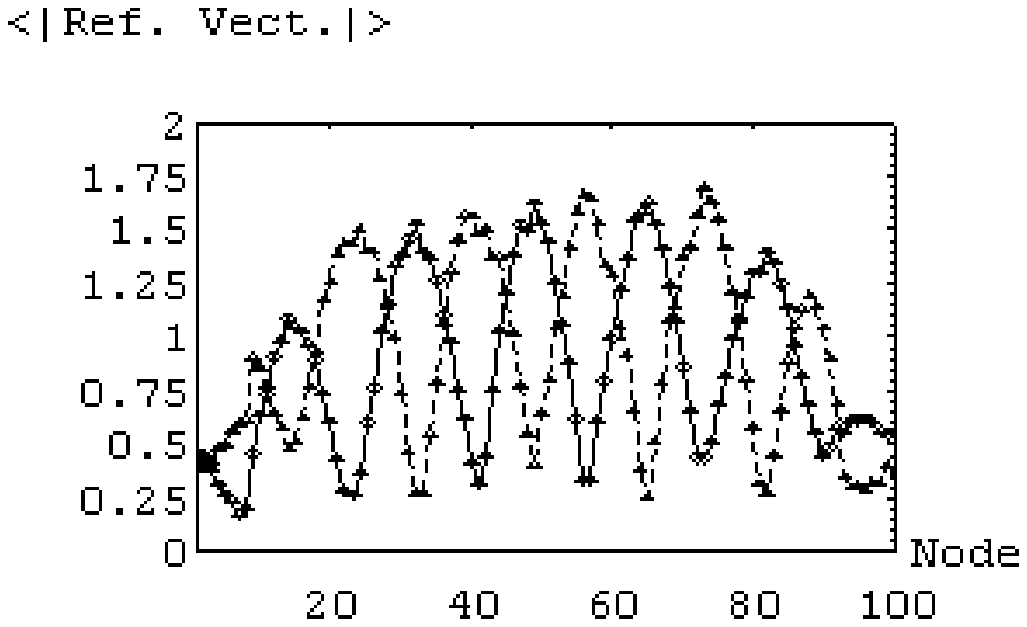}\caption{Dominance plots for a 1D simulation with 2 statistically independent
training set subspaces.}

\par\end{centering}

\end{figure}

The parameter values used were: $\left(m_{1},m_{2}\right)=(1,100)$,
$\left(i_{1},i_{2}\right)=\left(1,41\right)$, $\left(w_{1},w_{2}\right)=\left(1,21\right)$,
$\left(l_{1},l_{2}\right)=\left(1,15\right)$, $\kappa=0.3$, $\nu=0.1$,
$s=2$, $n=400$, $\varepsilon=0.002$. This value of $\kappa$ implies
$\frac{\kappa i_{2}}{2\pi}\simeq1.96$, so $\kappa$ is approximately
an integer multiple of $2\pi$, as required for an artefact-free simulation.
In figure \ref{Fig:Dominance1D} a plot of the 2 dominance curves
obtained after 3200 training updates is shown. %
This dominance plot clearly shows alternating regions where subspace
1 dominates and subspace 2 dominates. The width of the neighbourhood
function is 21, which is the same the period of the variations in
the dominance plots, i.e. within each set of adjacent 21 nodes half
the nodes are attached to subspace 1 and half to subspace 2. There
are boundary effects, but these are unimportant.

\appendix

\section{Normalisation of $\Pr\left(\mathbf{y}|\mathbf{x}\right)$}

\label{App:PostProbPMDNorm}

The normalisation of the expression for $\Pr\left(\mathbf{y}|\mathbf{x}\right)$
in equation \ref{Eq:PostProbPMD} may be demonstrated as follows:
\begin{eqnarray}
\sum_{\mathbf{y}=\mathbf{1}}^{\mathbf{m}}\Pr\left(\mathbf{y}|\mathbf{x}\right) & = & \frac{1}{M}\sum_{\mathbf{y}=\mathbf{1}}^{\mathbf{m}}Q\left(\mathbf{x}|\mathbf{y}\right)\sum_{\mathbf{y}^{\prime}\in\tilde{N}\left(\mathbf{y}\right)}\frac{1}{\sum_{\mathbf{y}^{\prime\prime}\in N\left(\mathbf{y}^{\prime}\right)}Q\left(\mathbf{x}|\mathbf{y}^{\prime\prime}\right)}\nonumber \\
 & = & \frac{1}{M}\sum_{\mathbf{y}^{\prime}=\mathbf{1}}^{\mathbf{m}}\sum_{\mathbf{y}\in N\left(\mathbf{y}^{\prime}\right)}Q\left(\mathbf{x}|\mathbf{y}\right)\frac{1}{\sum_{\mathbf{y}^{\prime\prime}\in N\left(\mathbf{y}^{\prime}\right)}Q\left(\mathbf{x}|\mathbf{y}^{\prime\prime}\right)}\nonumber \\
 & = & \frac{1}{M}\sum_{\mathbf{y}^{\prime}=\mathbf{1}}^{\mathbf{m}}1\nonumber \\
 & = & \frac{m_{1}m_{2}\,\cdots\, m_{d}}{M}\nonumber \\
 & = & 1\label{Eq:PostProbPMDNorm}\end{eqnarray}
 In the first step the order of the $\mathbf{y}$ and the $\mathbf{y}^{\prime}$
summations is interchanged using $\sum_{\mathbf{y}=\mathbf{1}}^{\mathbf{m}}\sum_{\mathbf{y}^{\prime}\in\tilde{N}\left(\mathbf{y}\right)}\left(\cdots\right)=\sum_{\mathbf{y}^{\prime}=\mathbf{1}}^{\mathbf{m}}\sum_{\mathbf{y}\in N\left(\mathbf{y}^{\prime}\right)}\left(\cdots\right)$,
in the second step the numerator and denominator of the summand cancel
out.

\section{Upper Bound for Multiple Firing Model}

\label{App:UpperBound}

It is possible to simplify equation \ref{Eq:ObjectiveMultiFire} by
using the following identity \begin{equation}
\mathbf{x}-\mathbf{x}^{\prime}\left(\mathbf{y}_{1},\mathbf{y}_{2},\cdots\mathbf{y}_{n}\right)\equiv\frac{1}{n}\sum_{i=1}^{n}\left(\mathbf{x}-\mathbf{x}^{\prime}\left(\mathbf{y}_{i}\right)\right)+\frac{1}{n}\sum_{i=1}^{n}\left(\mathbf{x}^{\prime}\left(\mathbf{y}_{i}\right)-\mathbf{x}^{\prime}\left(\mathbf{y}_{1},\mathbf{y}_{2},\cdots\mathbf{y}_{n}\right)\right)\label{Eq:DeltaSplit}\end{equation}
 Note that this holds for all choices of $\mathbf{x}^{\prime}\left(\mathbf{y}_{i}\right)$.
This allows the Euclidean distance to be expanded thus \begin{eqnarray}
 &  & \ \ \left\Vert \mathbf{x}-\mathbf{x}^{\prime}\left(\mathbf{y}_{1},\mathbf{y}_{2},\cdots\mathbf{y}_{n}\right)\right\Vert ^{2}\nonumber \\
\  & = & \frac{1}{n^{2}}\sum_{i=1}^{n}\left\Vert \mathbf{x}-\mathbf{x}^{\prime}\left(\mathbf{y}_{i}\right)\right\Vert ^{2}\nonumber \\
 &  & \ \ \ \ \ +\frac{1}{n^{2}}\sum_{\begin{array}{c}
i,j=1\\
i\neq j\end{array}}^{n}\left(\mathbf{x}-\mathbf{x}^{\prime}\left(\mathbf{y}_{i}\right)\right)\cdot\left(\mathbf{x}-\mathbf{x}^{\prime}\left(\mathbf{y}_{j}\right)\right)\\
 &  & \ \ \ \ \ \ -\frac{2}{n^{2}}\sum_{i,j=1}^{n}\left(\mathbf{x}-\mathbf{x}^{\prime}\left(\mathbf{y}_{i}\right)\right)\left(\mathbf{x}^{\prime}\left(\mathbf{y}_{j}\right)-\mathbf{x}^{\prime}\left(\mathbf{y}_{1},\mathbf{y}_{2},\cdots\mathbf{y}_{n}\right)\right)\nonumber \\
 &  & \ \ \ \ \ \ +\frac{1}{n^{2}}\left\Vert \sum_{i=1}^{n}\left(\mathbf{x}^{\prime}\left(\mathbf{y}_{i}\right)-\mathbf{x}^{\prime}\left(\mathbf{y}_{1},\mathbf{y}_{2},\cdots\mathbf{y}_{n}\right)\right)\right\Vert ^{2}\label{Eq:EuclideanSplit}\end{eqnarray}
 Each term of this expansion can be inserted into equation \ref{Eq:ObjectiveMultiFire}
to yield \begin{eqnarray}
\text{term 1} & = & \frac{1}{n}\sum_{\mathbf{y}=\mathbf{1}}^{\mathbf{m}}\int d\mathbf{x}\Pr\left(\mathbf{x}\right)\Pr\left(\mathbf{y}|\mathbf{x}\right)\left\Vert \mathbf{x}-\mathbf{x}^{\prime}\left(\mathbf{y}\right)\right\Vert ^{2}\nonumber \\
\text{term 2} & = & \frac{n-1}{n}\int d\mathbf{x}\Pr\left(\mathbf{x}\right)\sum_{\mathbf{y}_{1},\mathbf{y}_{2}=\mathbf{1}}^{\mathbf{m}}\Pr\left(\mathbf{y}_{1},\mathbf{y}_{2}|\mathbf{x}\right)\left(\mathbf{x}-\mathbf{x}^{\prime}\left(\mathbf{y}_{1}\right)\right)\left(\mathbf{x}-\mathbf{x}^{\prime}\left(\mathbf{y}_{2}\right)\right)\nonumber \\
\text{term 3} & = & -2\times\text{term 4}\nonumber \\
\text{term 4} & = & \sum_{\mathbf{y}_{1},\mathbf{y}_{2},\cdots\mathbf{y}_{n}=\mathbf{1}}^{\mathbf{m}}\Pr\left(\mathbf{y}_{1},\mathbf{y}_{2},\cdots\mathbf{y}_{n}\right)\left\Vert \begin{array}{c}
\mathbf{x}^{\prime}\left(\mathbf{y}_{1},\mathbf{y}_{2},\cdots\mathbf{y}_{n}\right)\\
-\frac{1}{n}\sum_{i=1}^{n}\mathbf{x}^{\prime}\left(\mathbf{y}_{i}\right)\end{array}\right\Vert ^{2}\end{eqnarray}
 $\Pr\left(\mathbf{y}_{1},\mathbf{y}_{2},\cdots\mathbf{y}_{n}|\mathbf{x}\right)$
has been assumed to be a symmetric function of $\left(\mathbf{y}_{1},\mathbf{y}_{2},\cdots\mathbf{y}_{n}\right)$
in the first two results, and the definition of $\mathbf{x}^{\prime}\left(\mathbf{y}_{1},\mathbf{y}_{2},\cdots\mathbf{y}_{n}\right)$
in equation \ref{Eq:RefVectMultiFire} has been used to obtain the
third result. These results allow $D$ in equation \ref{Eq:ObjectiveMultiFire}
to be expanded as $D=D_{1}+D_{2}-D_{3}$, where \begin{eqnarray}
D_{1} & \equiv & \frac{2}{n}\sum_{\mathbf{y}=\mathbf{1}}^{\mathbf{m}}\int d\mathbf{x}\Pr\left(\mathbf{x}\right)\Pr\left(\mathbf{y}|\mathbf{x}\right)\left\Vert \mathbf{x}-\mathbf{x}^{\prime}\left(\mathbf{y}\right)\right\Vert ^{2}\nonumber \\
D_{2} & \equiv & \frac{2\left(n-1\right)}{n}\int d\mathbf{x}\Pr\left(\mathbf{x}\right)\sum_{\mathbf{y}_{1},\mathbf{y}_{2}=\mathbf{1}}^{\mathbf{m}}\Pr\left(\mathbf{y}_{1},\mathbf{y}_{2}|\mathbf{x}\right)\left(\mathbf{x}-\mathbf{x}^{\prime}\left(\mathbf{y}_{1}\right)\right)\left(\mathbf{x}-\mathbf{x}^{\prime}\left(\mathbf{y}_{2}\right)\right)\nonumber \\
D_{3} & \equiv & 2\sum_{\mathbf{y}_{1},\mathbf{y}_{2},\cdots\mathbf{y}_{n}=\mathbf{1}}^{\mathbf{m}}\Pr\left(\mathbf{y}_{1},\mathbf{y}_{2},\cdots\mathbf{y}_{n}\right)\left\Vert \begin{array}{c}
\mathbf{x}^{\prime}\left(\mathbf{y}_{1},\mathbf{y}_{2},\cdots\mathbf{y}_{n}\right)\\
-\frac{1}{n}\sum_{i=1}^{n}\mathbf{x}^{\prime}\left(\mathbf{y}_{i}\right)\end{array}\right\Vert ^{2}\label{Eq:UpperBoundPiecesBis}\end{eqnarray}
 By noting that $D_{3}\geq0$, an upper bound for $D$ in the form
$D\leq D_{1}+D_{2}$ follows immediately from these results. Note
that $D_{1}\geq0$ whereas $D_{2}$ can have either sign. In the special
case where $\Pr\left(\mathbf{y}_{1},\mathbf{y}_{2}|\mathbf{x}\right)=\Pr\left(\mathbf{y}_{1}|\mathbf{x}\right)\Pr\left(\mathbf{y}_{1}|\mathbf{x}\right)$
(i.e. $\mathbf{y}_{1}$ and $\mathbf{y}_{2}$ are independent of each
other given that $x$ is known) $D_{2}$ reduces to \begin{equation}
D_{2}\equiv\frac{2\left(n-1\right)}{n}\int d\mathbf{x}\Pr\left(\mathbf{x}\right)\left\Vert \sum_{\mathbf{y}=\mathbf{1}}^{\mathbf{m}}\Pr\left(\mathbf{y}|\mathbf{x}\right)\left(\mathbf{x}-\mathbf{x}^{\prime}\left(\mathbf{y}\right)\right)\right\Vert ^{2}\end{equation}
 which is manifestly positive. This is the form of $D_{2}$ that is
used throughout this paper.

\section{Derivatives of the Objective Function}

\label{App:Derivatives}

$D_{1}$ and $D_{2}$ are as given in equation \ref{Eq:UpperBoundPieces},
i.e. it is assumed that $\Pr\left(\mathbf{y}_{1},\mathbf{y}_{2}|\mathbf{x}\right)=\Pr\left(\mathbf{y}_{2}|\mathbf{x}\right)\Pr\left(\mathbf{y}_{2}|\mathbf{x}\right)$
and $\Pr\left(\mathbf{y}|\mathbf{x}\right)$ has the scalable form
given in equation \ref{Eq:PostProbPMD}. Define a compact matrix notation
as follows \begin{eqnarray}
\ \begin{array}{ll}
L_{\mathbf{y},\mathbf{y}^{\prime}}\equiv\Pr\left(\mathbf{y^{\prime}|y}\right) & P_{\mathbf{y,y}^{\prime}}\equiv\Pr\left(\mathbf{y^{\prime}|x;y}\right)\\
p_{\mathbf{y}}\equiv\sum_{\mathbf{y}^{\prime}\in\tilde{N}\left(\mathbf{y}\right)}P_{\mathbf{y^{\prime},y}} & \left(L^{T}p\right)_{\mathbf{y}}\equiv\sum_{\mathbf{y}^{\prime}=\mathbf{1}}^{\mathbf{m}}L_{\mathbf{y}^{\prime}\mathbf{,y}}\, p_{\mathbf{y}^{\prime}}\\
\mathbf{d}_{\mathbf{y}}\equiv\mathbf{x}-\mathbf{x}^{\prime}\left(\mathbf{y}\right) & \left(L\,\mathbf{d}\right)_{\mathbf{y}}\equiv\sum_{\mathbf{y}^{\prime}=\mathbf{1}}^{\mathbf{m}}L_{\mathbf{y},\mathbf{y}^{\prime}}\,\mathbf{d}_{\mathbf{y}^{\prime}}\\
\left(P\, L\,\mathbf{d}\right)_{\mathbf{y}}\equiv\sum_{\mathbf{y}^{\prime}\in N\left(\mathbf{y}\right)}P_{\mathbf{y},\mathbf{y}^{\prime}}\left(L\,\mathbf{d}\right)_{\mathbf{y}^{\prime}} & \left(P^{T}P\, L\,\mathbf{d}\right)_{\mathbf{y}}\equiv\sum_{\mathbf{y}^{\prime}\in\tilde{N}\left(\mathbf{y}\right)}P_{\mathbf{y^{\prime},y}}\left(P\, L\,\mathbf{d}\right)_{\mathbf{y}^{\prime}}\\
e_{\mathbf{y}}\equiv\left\Vert \mathbf{x}-\mathbf{x}^{\prime}\left(\mathbf{y}\right)\right\Vert ^{2} & \left(L\, e\right)_{\mathbf{y}}\equiv\sum_{\mathbf{y}^{\prime}=\mathbf{1}}^{\mathbf{m}}L_{\mathbf{y},\mathbf{y}^{\prime}}\, e_{\mathbf{y}^{\prime}}\\
\left(P\, L\, e\right)_{\mathbf{y}}\equiv\sum_{\mathbf{y}^{\prime}\in N\left(\mathbf{y}\right)}P_{\mathbf{y,y}^{\prime}}\left(L\, e\right)_{\mathbf{y}^{\prime}} & \left(P^{T}P\, L\, e\right)_{\mathbf{y}}\equiv\sum_{\mathbf{y}^{\prime}\in\tilde{N}\left(\mathbf{y}\right)}P_{\mathbf{y^{\prime},y}}\left(P\, L\, e\right)_{\mathbf{y}^{\prime}}\\
\mathbf{\bar{d}\equiv}\sum_{\mathbf{y}=\mathbf{1}}^{\mathbf{m}}\left(L^{T}p\right)_{\mathbf{y}}\mathbf{d}_{\mathbf{y}} & \text{or }\mathbf{\bar{d}\equiv}\sum_{\mathbf{y}=\mathbf{1}}^{\mathbf{m}}\left(P\, L\,\mathbf{d}\right)_{\mathbf{y}}\end{array}\label{Eq:MatrixNotation}\end{eqnarray}
 Using this matrix notation, the functions $\mathbf{f}_{1}\left(\mathbf{x},\mathbf{y}\right)$,
$\,\mathbf{f}_{2}\left(\mathbf{x},\mathbf{y}\right)$, $g_{1}\left(\mathbf{x},\mathbf{y}\right)$,
and $g_{2}\left(\mathbf{x},\mathbf{y}\right)$ may be defined as \begin{eqnarray}
\mathbf{f}_{1}\left(\mathbf{x},\mathbf{y}\right) & \equiv & \left(L^{T}p\right)_{\mathbf{y}}\mathbf{d}_{\mathbf{y}}\label{Eq:F1}\\
\mathbf{f}_{2}\left(\mathbf{x},\mathbf{y}\right) & \equiv & \left(L^{T}p\right)_{\mathbf{y}}\mathbf{\bar{d}}\label{Eq:F2}\\
g_{1}\left(\mathbf{x},\mathbf{y}\right) & \equiv & p_{\mathbf{y}}\left(L\, e\right)_{\mathbf{y}}-\left(P^{T}P\, L\, e\right)_{\mathbf{y}}\label{Eq:G1}\\
g_{2}\left(\mathbf{x},\mathbf{y}\right) & \equiv & \left(p_{\mathbf{y}}\left(L\,\mathbf{d}\right)_{\mathbf{y}}-\left(P^{T}P\, L\,\mathbf{d}\right)_{\mathbf{y}}\right)\cdot\mathbf{\bar{d}}\label{Eq:G2}\end{eqnarray}
 The variation of $\Pr\left(\mathbf{y}|\mathbf{x};\mathbf{y}^{\prime}\right)$
is then given by \begin{eqnarray}
\delta\Pr\left(\mathbf{y}|\mathbf{x};\mathbf{y}^{\prime}\right) & = & \Pr\left(\mathbf{y}|\mathbf{x};\mathbf{y}^{\prime}\right)\left(\begin{array}{c}
\delta\log Q\left(\mathbf{x}|\mathbf{y}\right)\\
-\sum_{\mathbf{y}^{\prime\prime}\in N\left(\mathbf{y}^{\prime}\right)}\Pr\left(\mathbf{y}^{\prime\prime}|\mathbf{x};\mathbf{y}^{\prime}\right)\delta\log Q\left(\mathbf{x}|\mathbf{y}^{\prime\prime}\right)\end{array}\right)\nonumber \\
\  & = & \Pr\left(\mathbf{y}|\mathbf{x};\mathbf{y}^{\prime}\right)\sum_{\mathbf{y}^{\prime\prime}\in N\left(\mathbf{y}^{\prime}\right)}\delta\log Q\left(\mathbf{x}|\mathbf{y}^{\prime\prime}\right)\left(\delta_{\mathbf{y}^{\prime\prime},\mathbf{y}}-\Pr\left(\mathbf{y}^{\prime\prime}|\mathbf{x};\mathbf{y}^{\prime}\right)\right)\nonumber \\
\  & = & P_{\mathbf{y}^{\prime},\mathbf{y}}\sum_{\mathbf{y}^{\prime\prime}\in N\left(\mathbf{y}^{\prime}\right)}\delta\log Q\left(\mathbf{x}|\mathbf{y}^{\prime\prime}\right)\left(\delta_{\mathbf{y}^{\prime\prime},\mathbf{y}}-P_{\mathbf{y}^{\prime},\mathbf{y}^{\prime\prime}}\right)\label{Eq:DeltaPostProbMD}\end{eqnarray}
 In order to rearrange the expressions to ensure that only a single
dummy index is required at every stage of evaluation of the sums it
will be necessary to use the result \begin{equation}
\sum_{\mathbf{y}=\mathbf{1}}^{\mathbf{m}}\sum_{\mathbf{y}^{\prime}\in N\left(\mathbf{y}\right)}=\sum_{\mathbf{y}^{\prime}=\mathbf{1}}^{\mathbf{m}}\sum_{\mathbf{y}\in\tilde{N}\left(\mathbf{y}^{\prime}\right)}\label{Eq:ExchangeSum}\end{equation}

\subsection{Calculate $\frac{\partial D_{1}}{\partial\mathbf{x}^{\prime}\left(\mathbf{y}\right)}$.}

The derivative is given by \begin{equation}
\frac{\partial D_{1}}{\partial\mathbf{x}^{\prime}\left(\mathbf{y}\right)}=-\frac{4}{n\, M}\int d\mathbf{x}\Pr\left(\mathbf{x}\right)\sum_{\mathbf{y}^{\prime}=\mathbf{1}}^{\mathbf{m}}\Pr\left(\mathbf{y}|\mathbf{y}^{\prime}\right)\sum_{\mathbf{y}^{\prime\prime}\in\tilde{N}\left(\mathbf{y}^{\prime}\right)}\Pr\left(\mathbf{y}^{\prime}|\mathbf{x};\mathbf{y}^{\prime\prime}\right)\left(\mathbf{x}-\mathbf{x}^{\prime}\left(\mathbf{y}\right)\right)\end{equation}
 Use matrix notation to write this as \begin{equation}
\frac{\partial D_{1}}{\partial\mathbf{x}^{\prime}\left(\mathbf{y}\right)}=-\frac{4}{n\, M}\int d\mathbf{x}\Pr\left(\mathbf{x}\right)\sum_{\mathbf{y}^{\prime}=\mathbf{1}}^{\mathbf{m}}L_{\mathbf{y}^{\prime},\mathbf{y}}\sum_{\mathbf{y}^{\prime\prime}\in\tilde{N}\left(\mathbf{y}^{\prime}\right)}P_{\mathbf{y}^{\prime\prime}\mathbf{,y}^{\prime}}\,\mathbf{d}_{\mathbf{y}}\end{equation}
 Finally remove the explicit summations to obtain the required result
\begin{eqnarray}
\frac{\partial D_{1}}{\partial\mathbf{x}^{\prime}\left(\mathbf{y}\right)} & = & -\frac{4}{n\, M}\int d\mathbf{x}\Pr\left(\mathbf{x}\right)\left(L^{T}p\right)_{\mathbf{y}}\mathbf{d}_{\mathbf{y}}\label{Eq:DD1DRefVect}\\
\  & = & -\frac{4}{n\, M}\int d\mathbf{x}\Pr\left(\mathbf{x}\right)\,\mathbf{f}_{1}\left(\mathbf{x},\mathbf{y}\right)\nonumber \end{eqnarray}

\subsection{Calculate $\frac{\partial D_{2}}{\partial\mathbf{x}^{\prime}\left(\mathbf{y}\right)}.$}

The derivative is given by \begin{eqnarray}
\frac{\partial D_{2}}{\partial\mathbf{x}^{\prime}\left(\mathbf{y}\right)} & = & -\frac{4\left(n-1\right)}{n\, M^{2}}\int d\mathbf{x}\Pr\left(\mathbf{x}\right)\left(\sum_{\mathbf{y}^{\prime}=\mathbf{1}}^{\mathbf{m}}\Pr\left(\mathbf{y}|\mathbf{y}^{\prime}\right)\sum_{\mathbf{y}^{\prime\prime}\in\tilde{N}\left(\mathbf{y}^{\prime}\right)}\Pr\left(\mathbf{y}^{\prime}|\mathbf{x};\mathbf{y}^{\prime\prime}\right)\right)\nonumber \\
 &  & \ \times\left(\sum_{\mathbf{\bar{y}}=\mathbf{1}}^{\mathbf{m}}\sum_{\mathbf{y}^{\prime}=\mathbf{1}}^{\mathbf{m}}\Pr\left(\mathbf{\bar{y}}|\mathbf{y}^{\prime}\right)\sum_{\mathbf{y}^{\prime\prime}\in\tilde{N}\left(\mathbf{y}^{\prime}\right)}\Pr\left(\mathbf{y}^{\prime}|\mathbf{x};\mathbf{y}^{\prime\prime}\right)\left(\mathbf{x}-\mathbf{x}^{\prime}\left(\mathbf{\bar{y}}\right)\right)\right)\end{eqnarray}
 Use matrix notation to write this as \begin{eqnarray}
\frac{\partial D_{2}}{\partial\mathbf{x}^{\prime}\left(\mathbf{y}\right)} & = & -\frac{4\left(n-1\right)}{n\, M^{2}}\int d\mathbf{x}\Pr\left(\mathbf{x}\right)\left(\sum_{\mathbf{y}^{\prime}=\mathbf{1}}^{\mathbf{m}}L_{\mathbf{y}^{\prime},\mathbf{y}}\sum_{\mathbf{y}^{\prime\prime}\in\tilde{N}\left(\mathbf{y}^{\prime}\right)}P_{\mathbf{y}^{\prime\prime}\mathbf{,y}^{\prime}}\right)\nonumber \\
 &  & \ \times\left(\sum_{\mathbf{\bar{y}}=\mathbf{1}}^{\mathbf{m}}\sum_{\mathbf{y}^{\prime}=\mathbf{1}}^{\mathbf{m}}L_{\mathbf{y}^{\prime},\mathbf{\bar{y}}}\sum_{\mathbf{y}^{\prime\prime}\in\tilde{N}\left(\mathbf{y}^{\prime}\right)}P_{\mathbf{y}^{\prime\prime}\mathbf{,y}^{\prime}}\mathbf{d}_{\mathbf{\bar{y}}}\right)\end{eqnarray}
 Finally remove the explicit summations to obtain the required result
\begin{eqnarray}
\frac{\partial D_{2}}{\partial\mathbf{x}^{\prime}\left(\mathbf{y}\right)} & = & -\frac{4\left(n-1\right)}{n\, M^{2}}\int d\mathbf{x}\Pr\left(\mathbf{x}\right)\left(L^{T}p\right)_{\mathbf{y}}\mathbf{\bar{d}}\nonumber \\
\  & = & -\frac{4\left(n-1\right)}{n\, M^{2}}\int d\mathbf{x}\Pr\left(\mathbf{x}\right)\,\mathbf{f}_{2}\left(\mathbf{x},\mathbf{y}\right)\label{Eq:DD2DRefVect}\end{eqnarray}

\subsection{Calculate $\frac{\delta D_{1}}{\delta\log Q\left(\mathbf{x}|\mathbf{y}\right)}$}

The differential is given by \begin{equation}
\delta D_{1}=\frac{2}{n\, M}\sum_{\mathbf{y}=1}^{\mathbf{m}}\int d\mathbf{x}\Pr\left(\mathbf{x}\right)\sum_{\mathbf{y}^{\prime}=\mathbf{1}}^{\mathbf{m}}\Pr\left(\mathbf{y}|\mathbf{y}^{\prime}\right)\sum_{\mathbf{y}^{\prime\prime}\in\tilde{N}\left(\mathbf{y}^{\prime}\right)}\delta\Pr\left(\mathbf{y}^{\prime}|\mathbf{x};\mathbf{y}^{\prime\prime}\right)\left\Vert \mathbf{x}-\mathbf{x}^{\prime}\left(\mathbf{y}\right)\right\Vert ^{2}\end{equation}
 Use matrix notation to write this as \begin{equation}
\delta D_{1}=\frac{2}{n\, M}\sum_{\mathbf{y}=\mathbf{1}}^{\mathbf{m}}\int d\mathbf{x}\Pr\left(\mathbf{x}\right)\left(\begin{array}{c}
\sum_{\mathbf{y}^{\prime}=\mathbf{1}}^{\mathbf{m}}L_{\mathbf{y}^{\prime}\mathbf{,y}}\sum_{\mathbf{y}^{\prime\prime}\in\tilde{N}\left(\mathbf{y}^{\prime}\right)}P_{\mathbf{y}^{\prime\prime},\mathbf{y}^{\prime}}\\
\times\sum_{\mathbf{y}^{\prime\prime\prime}\in N\left(\mathbf{y}^{\prime\prime}\right)}\delta\log Q\left(\mathbf{x}|\mathbf{y}^{\prime\prime\prime}\right)\left(\delta_{\mathbf{y}^{\prime\prime\prime},\mathbf{y}^{\prime}}-P_{\mathbf{y}^{\prime\prime},\mathbf{y}^{\prime\prime\prime}}\right)\end{array}\right)e_{\mathbf{y}}\end{equation}
 Reorder the summations to obtain \begin{equation}
\delta D_{1}=\frac{2}{n\, M}\int d\mathbf{x}\Pr\left(\mathbf{x}\right)\left(\begin{array}{c}
\sum_{\mathbf{y}^{\prime\prime\prime}=\mathbf{1}}^{\mathbf{m}}\delta\log Q\left(\mathbf{x}|\mathbf{y}^{\prime\prime\prime}\right)\sum_{\mathbf{y}^{\prime\prime}\in\tilde{N}\left(\mathbf{y}^{\prime\prime\prime}\right)}\\
\times\left(\begin{array}{c}
\sum_{\mathbf{y}^{\prime}\in N\left(\mathbf{y}^{\prime\prime}\right)}\delta_{\mathbf{y}^{\prime\prime\prime},\mathbf{y}^{\prime}}P_{\mathbf{y}^{\prime\prime},\mathbf{y}^{\prime}}\\
-P_{\mathbf{y}^{\prime\prime},\mathbf{y}^{\prime\prime\prime}}\sum_{\mathbf{y}^{\prime}\in N\left(\mathbf{y}^{\prime\prime}\right)}P_{\mathbf{y}^{\prime\prime},\mathbf{y}^{\prime}}\end{array}\right)\sum_{\mathbf{y}=1}^{\mathbf{m}}L_{\mathbf{y}^{\prime}\mathbf{,y}}\, e_{\mathbf{y}}\end{array}\right)\end{equation}
 Relabel the indices and evaluate the sum over the Kronecker delta
to obtain \begin{eqnarray}
\delta D_{1} & = & \frac{2}{n\, M}\sum_{\mathbf{y}=\mathbf{1}}^{\mathbf{m}}\int d\mathbf{x}\Pr\left(\mathbf{x}\right)\delta\log Q\left(\mathbf{x}|\mathbf{y}\right)\\
 &  & \times\left(\begin{array}{c}
\left(\sum_{\mathbf{y}^{\prime}\in\tilde{N}\left(\mathbf{y}\right)}P_{\mathbf{y}^{\prime},\mathbf{y}}\right)\left(\sum_{\mathbf{y}^{\prime}=\mathbf{1}}^{\mathbf{m}}L_{\mathbf{y,y}^{\prime}}\, e_{\mathbf{y}^{\prime}}\right)\\
-\left(\sum_{\mathbf{y}^{\prime\prime\prime}\in\tilde{N}\left(\mathbf{y}\right)}P_{\mathbf{y}^{\prime\prime\prime},\mathbf{y}}\sum_{\mathbf{y}^{\prime\prime}\in N\left(\mathbf{y}^{\prime\prime\prime}\right)}P_{\mathbf{y}^{\prime\prime\prime},\mathbf{y}^{\prime\prime}}\sum_{\mathbf{y}^{\prime}=\mathbf{1}}^{\mathbf{m}}L_{\mathbf{y}^{\prime\prime},\mathbf{y}^{\prime}}\, e_{\mathbf{y}^{\prime}}\right)\end{array}\right)\nonumber \end{eqnarray}
 Finally remove the explicit summations to obtain the required result
\begin{eqnarray}
\delta D_{1} & = & \frac{2}{n\, M}\sum_{\mathbf{y}=\mathbf{1}}^{\mathbf{m}}\int d\mathbf{x}\Pr\left(\mathbf{x}\right)\delta\log Q\left(\mathbf{x}|\mathbf{y}\right)\left(p_{\mathbf{y}}\left(L\, e\right)_{\mathbf{y}}-(P^{T}P\, L\, e)_{\mathbf{y}}\right)\nonumber \\
 & = & \frac{2}{n\, M}\sum_{\mathbf{y}=\mathbf{1}}^{\mathbf{m}}\int d\mathbf{x}\Pr\left(\mathbf{x}\right)g_{1}\left(\mathbf{x},\mathbf{y}\right)\,\delta\log Q\left(\mathbf{x}|\mathbf{y}\right)\end{eqnarray}

\subsection{Calculate $\frac{\delta D_{2}}{\delta\log Q\left(\mathbf{x}|\mathbf{y}\right)}$}

The differential is given by \begin{eqnarray}
\delta D_{2} & = & \frac{4\left(n-1\right)}{n\, M^{2}}\int d\mathbf{x}\Pr\left(\mathbf{x}\right)\nonumber \\
 &  & \times\left(\sum_{\mathbf{y}=\mathbf{1}}^{\mathbf{m}}\sum_{\mathbf{y}^{\prime}=\mathbf{1}}^{\mathbf{m}}\Pr\left(\mathbf{y}|\mathbf{y}^{\prime}\right)\sum_{\mathbf{y}^{\prime\prime}\in\tilde{N}\left(\mathbf{y}^{\prime}\right)}\Pr\left(\mathbf{y}^{\prime}|\mathbf{x};\mathbf{y}^{\prime\prime}\right)\left(\mathbf{x}-\mathbf{x}^{\prime}\left(\mathbf{y}\right)\right)\right)\nonumber \\
 &  & \ \cdot\left(\sum_{\mathbf{y}=\mathbf{1}}^{\mathbf{m}}\sum_{\mathbf{y}^{\prime}=\mathbf{1}}^{\mathbf{m}}\Pr\left(\mathbf{y}|\mathbf{y}^{\prime}\right)\sum_{\mathbf{y}^{\prime\prime}\in\tilde{N}\left(\mathbf{y}^{\prime}\right)}\delta\Pr\left(\mathbf{y}^{\prime}|\mathbf{x};\mathbf{y}^{\prime\prime}\right)\left(\mathbf{x}-\mathbf{x}^{\prime}\left(\mathbf{y}\right)\right)\right)\end{eqnarray}
 Use matrix notation to write this as \begin{eqnarray}
\delta D_{2} & = & \frac{4\left(n-1\right)}{n\, M^{2}}\int d\mathbf{x}\Pr\left(\mathbf{x}\right)\left(\sum_{\mathbf{y}=\mathbf{1}}^{\mathbf{m}}\sum_{\mathbf{y}^{\prime}=\mathbf{1}}^{\mathbf{m}}L_{\mathbf{y}^{\prime}\mathbf{,y}}\sum_{\mathbf{y}^{\prime\prime}\in\tilde{N}\left(\mathbf{y}^{\prime}\right)}P_{\mathbf{y}^{\prime\prime},\mathbf{y}^{\prime}}\,\mathbf{d}_{\mathbf{y}}\right)\\
 &  & \ \cdot\left(\sum_{\mathbf{y}=\mathbf{1}}^{\mathbf{m}}\begin{array}{c}
\sum_{\mathbf{y}^{\prime}=\mathbf{1}}^{\mathbf{m}}L_{\mathbf{y}^{\prime}\mathbf{,y}}\sum_{\mathbf{y}^{\prime\prime}\in\tilde{N}\left(\mathbf{y}^{\prime}\right)}P_{\mathbf{y}^{\prime\prime},\mathbf{y}^{\prime}}\\
\times\sum_{\mathbf{y}^{\prime\prime\prime}\in N\left(\mathbf{y}^{\prime\prime}\right)}\delta\log Q\left(\mathbf{x}|\mathbf{y}^{\prime\prime\prime}\right)\left(\delta_{\mathbf{y}^{\prime\prime\prime},\mathbf{y}^{\prime}}-P_{\mathbf{y}^{\prime\prime},\mathbf{y}^{\prime\prime\prime}}\right)\end{array}\mathbf{d}_{\mathbf{y}}\right)\nonumber \end{eqnarray}
 Reorder the summations to obtain \begin{eqnarray}
\delta D_{2} & = & \frac{4\left(n-1\right)}{n\, M^{2}}\int d\mathbf{x}\Pr\left(\mathbf{x}\right)\left(\sum_{\mathbf{y}=\mathbf{1}}^{\mathbf{m}}\sum_{\mathbf{y}^{\prime}=\mathbf{1}}^{\mathbf{m}}L_{\mathbf{y}^{\prime}\mathbf{,y}}\sum_{\mathbf{y}^{\prime\prime}\in\tilde{N}\left(\mathbf{y}^{\prime}\right)}P_{\mathbf{y}^{\prime\prime},\mathbf{y}^{\prime}}\,\mathbf{d}_{\mathbf{y}}\right)\nonumber \\
 &  & \ \cdot\left(\begin{array}{c}
\sum_{\mathbf{y}^{\prime\prime\prime}=\mathbf{1}}^{\mathbf{m}}\delta\log Q\left(\mathbf{x}|\mathbf{y}^{\prime\prime\prime}\right)\sum_{\mathbf{y}^{\prime\prime}\in\tilde{N}\left(\mathbf{y}^{\prime\prime\prime}\right)}\\
\times\left(\begin{array}{c}
\sum_{\mathbf{y}^{\prime}\in N\left(\mathbf{y}^{\prime\prime}\right)}\delta_{\mathbf{y}^{\prime\prime\prime},\mathbf{y}^{\prime}}P_{\mathbf{y}^{\prime\prime},\mathbf{y}^{\prime}}\\
-P_{\mathbf{y}^{\prime\prime},\mathbf{y}^{\prime\prime\prime}}\sum_{\mathbf{y}^{\prime}\in N\left(\mathbf{y}^{\prime\prime}\right)}P_{\mathbf{y}^{\prime\prime},\mathbf{y}^{\prime}}\end{array}\right)\sum_{\mathbf{y}=\mathbf{1}}^{\mathbf{m}}L_{\mathbf{y}^{\prime}\mathbf{,y}}\,\mathbf{d}_{\mathbf{y}}\end{array}\right)\end{eqnarray}
 Relabel the indices and evaluate the sum over the Kronecker delta
to obtain \begin{eqnarray}
\delta D_{2} & = & \frac{4\left(n-1\right)}{n\, M^{2}}\sum_{\mathbf{y}=\mathbf{1}}^{\mathbf{m}}\int d\mathbf{x}\Pr\left(\mathbf{x}\right)\delta\log Q\left(\mathbf{x}|\mathbf{y}\right)\nonumber \\
 &  & \times\left(\begin{array}{c}
\left(\sum_{\mathbf{y}^{\prime}\in\tilde{N}\left(\mathbf{y}\right)}P_{\mathbf{y}^{\prime},\mathbf{y}}\right)\left(\sum_{\mathbf{y}^{\prime}=\mathbf{1}}^{\mathbf{m}}L_{\mathbf{y,y}^{\prime}}\,\mathbf{d}_{\mathbf{y}^{\prime}}\right)\\
-\left(\sum_{\mathbf{y}^{\prime\prime\prime}\in\tilde{N}\left(\mathbf{y}\right)}P_{\mathbf{y}^{\prime\prime\prime},\mathbf{y}}\sum_{\mathbf{y}^{\prime\prime}\in N\left(\mathbf{y}^{\prime\prime\prime}\right)}P_{\mathbf{y}^{\prime\prime\prime},\mathbf{y}^{\prime\prime}}\sum_{\mathbf{y}^{\prime}=\mathbf{1}}^{\mathbf{m}}L_{\mathbf{y}^{\prime\prime}\mathbf{,y}^{\prime}}\,\mathbf{d}_{\mathbf{y}^{\prime}}\right)\end{array}\right)\nonumber \\
 &  & \cdot\left(\sum_{\mathbf{y}=\mathbf{1}}^{\mathbf{m}}\sum_{\mathbf{y}^{\prime}=\mathbf{1}}^{\mathbf{m}}L_{\mathbf{y}^{\prime}\mathbf{,y}}\sum_{\mathbf{y}^{\prime\prime}\in\tilde{N}\left(\mathbf{y}^{\prime}\right)}P_{\mathbf{y}^{\prime\prime},\mathbf{y}^{\prime}}\,\mathbf{d}_{\mathbf{y}}\right)\end{eqnarray}
 Finally remove the explicit summations to obtain the required result
\begin{eqnarray}
\delta D_{2} & = & \frac{4\left(n-1\right)}{n\, M^{2}}\sum_{\mathbf{y}=\mathbf{1}}^{\mathbf{m}}\int d\mathbf{x}\Pr\left(\mathbf{x}\right)\delta\log Q\left(\mathbf{x}|\mathbf{y}\right)\nonumber \\
 &  & \times\left(p_{\mathbf{y}}\left(L\,\mathbf{d}\right)_{\mathbf{y}}-\left(P^{T}P\, L\,\mathbf{d}\right)_{y}\right)\cdot\mathbf{\bar{d}}\nonumber \\
 & = & \frac{4\left(n-1\right)}{n\, M^{2}}\sum_{\mathbf{y}=\mathbf{1}}^{\mathbf{m}}\int d\mathbf{x}\Pr\left(\mathbf{x}\right)g_{2}\left(\mathbf{x},\mathbf{y}\right)\,\delta\log Q\left(\mathbf{x}|\mathbf{y}\right)\end{eqnarray}

\section{Expression for $D_{1}+D_{2}$ in Terms of $\mathbf{x}^{\prime}\left(c\right)$}

\label{App:D1D2RefVect}

From equation \ref{Eq:UpperBoundPieces} $D_{1}+D_{2}$ can be written
as \begin{eqnarray}
D_{1}+D_{2} & = & \frac{2}{n}\int d\mathbf{x}\Pr\left(\mathbf{x}\right)\sum_{\mathbf{y}=\mathbf{1}}^{\mathbf{m}}\Pr\left(\mathbf{y}|\mathbf{x}\right)\left(\left\Vert \mathbf{x}^{\prime}\left(\mathbf{y}\right)\right\Vert ^{2}-2\mathbf{x\cdot x}^{\prime}\left(\mathbf{y}\right)\right)\nonumber \\
 &  & \ \ \ +\frac{2\left(n-1\right)}{n}\int d\mathbf{x}\Pr\left(\mathbf{x}\right)\nonumber \\
 &  & \ \times\left(\begin{array}{c}
\left\Vert \sum_{\mathbf{y}=\mathbf{1}}^{\mathbf{m}}\Pr\left(\mathbf{y}|\mathbf{x}\right)\mathbf{x}^{\prime}\left(\mathbf{y}\right)\right\Vert ^{2}\\
-2\left(\sum_{\mathbf{y}=\mathbf{1}}^{\mathbf{m}}\Pr\left(\mathbf{y}|\mathbf{x}\right)\mathbf{x}^{\prime}\left(\mathbf{y}\right)\right)\cdot\left(\sum_{\mathbf{y}=\mathbf{1}}^{\mathbf{m}}\Pr\left(\mathbf{y}|\mathbf{x}\right)\mathbf{x}\right)\end{array}\right)\label{Eq:UpperBoundRefVectPieces}\\
 &  & \ \ \ +\text{constant}\nonumber \end{eqnarray}
 where the constant terms do not depend on $\mathbf{x}^{\prime}\left(\mathbf{y}\right)$.
However, from equation \ref{Eq:UpperBoundPieces} the derivative $\frac{\partial\left(D_{1}+D_{2}\right)}{\partial\mathbf{x}^{\prime}\left(\mathbf{y}\right)}$
can be written as \begin{eqnarray}
\frac{\partial\left(D_{1}+D_{2}\right)}{\partial\mathbf{x}^{\prime}\left(\mathbf{y}\right)} & = & -\frac{4}{n}\int d\mathbf{x}\Pr\left(\mathbf{x}\right)\Pr\left(\mathbf{y|x}\right)\nonumber \\
 &  & \ \times\left(\begin{array}{c}
\mathbf{x-x}^{\prime}\left(\mathbf{y}\right)\\
\\+\left(n-1\right)\sum_{\mathbf{y}^{\prime}=\mathbf{1}}^{\mathbf{m}}\Pr\left(\mathbf{y}^{\prime}\mathbf{|x}\right)\left(\mathbf{x-x}^{\prime}\left(\mathbf{y}^{\prime}\right)\right)\end{array}\right)\end{eqnarray}
 Using Bayes' theorem the stationarity condition $\frac{\partial\left(D_{1}+D_{2}\right)}{\partial\mathbf{x}^{\prime}\left(\mathbf{y}\right)}=0$
yields a matrix equation for the $\mathbf{x}^{\prime}\left(\mathbf{y}\right)$\begin{equation}
n\int d\mathbf{x}\Pr\left(\mathbf{x|y}\right)\,\mathbf{x=}\left(n-1\right)\sum_{\mathbf{y}^{\prime}=\mathbf{1}}^{\mathbf{m}}\left(\int d\mathbf{x}\Pr\left(\mathbf{x}|\mathbf{y}\right)\Pr\left(\mathbf{y}^{\prime}|\mathbf{x}\right)\right)\mathbf{x}^{\prime}\left(\mathbf{y}^{\prime}\right)+\mathbf{x}^{\prime}\left(\mathbf{y}\right)\end{equation}
 which may then be used to replace all instances of $\mathbf{x}$
in equation \ref{Eq:UpperBoundRefVectPieces}. This yields the result
\begin{eqnarray}
D_{1}+D_{2} & = & -\frac{2}{n}\int d\mathbf{x}\Pr\left(\mathbf{x}\right)\sum_{\mathbf{y}=\mathbf{1}}^{\mathbf{m}}\Pr\left(\mathbf{y}|\mathbf{x}\right)\left\Vert \mathbf{x}^{\prime}\left(\mathbf{y}\right)\right\Vert ^{2}\nonumber \\
 &  & \ \ \ -\frac{2\left(n-1\right)}{n}\int d\mathbf{x}\Pr\left(\mathbf{x}\right)\left\Vert \sum_{\mathbf{y}=\mathbf{1}}^{\mathbf{m}}\Pr\left(\mathbf{y}|\mathbf{x}\right)\mathbf{x}^{\prime}\left(\mathbf{y}\right)\right\Vert ^{2}\\
 &  & \ \ \ +\text{constant}\nonumber \end{eqnarray}

\section{Comparison of $D_{1}+D_{2}$ for Different Types of Optima}

\label{App:D1D2Compare}

In order to compare the value of $D_{1}+D_{2}$ that is obtained when
different types of supposedly optimum configurations of the threshold
functions $Q\left(\mathbf{x|y}\right)$ are tried, the $\mathbf{x}^{\prime}\left(\mathbf{y}\right)$
that solves $\frac{\partial\left(D_{1}+D_{2}\right)}{\partial\mathbf{x}^{\prime}\left(\mathbf{y}\right)}=0$
(see appendix \ref{App:D1D2RefVect})\ must be inserted into the
expression for $D_{1}+D_{2}$. In the following derivations the constant
term is omitted, and the definition $R_{M}\equiv\left\Vert \int d\mathbf{x}\Pr\left(\mathbf{x|}y\right)\,\mathbf{x}\right\Vert ^{2}$
(see equation \ref{Eq:RadiusSquared}) has been used.

\subsection{Type 1 Optimum: all the nodes are attached one subspace}

In equation \ref{Eq:D1D2RefVect} $D_{1}+D_{2}$ becomes \begin{eqnarray}
D_{1}+D_{2} & = & -\frac{2}{n}\int d\mathbf{x}\Pr\left(\mathbf{x}\right)\sum_{y=1}^{M}\Pr\left(y|\mathbf{x}\right)\left\Vert \left(\int d\mathbf{x}_{1}\Pr\left(\mathbf{x}_{1}|y\right)\,\mathbf{x}_{1},0\right)\right\Vert ^{2}\nonumber \\
 &  & \ -\frac{2\left(n-1\right)}{n}\int d\mathbf{x}\Pr\left(\mathbf{x}\right)\left\Vert \sum_{y=1}^{M}\Pr\left(y|\mathbf{x}\right)\left(\int d\mathbf{x}_{1}\Pr\left(\mathbf{x}_{1}|y\right)\,\mathbf{x}_{1},0\right)\right\Vert ^{2}\nonumber \\
\  & = & -\frac{2}{n}R_{M}-\frac{2\left(n-1\right)}{n}R_{M}\nonumber \\
\  & = & -2R_{M}\end{eqnarray}

\subsection{Type 2 Optimum:\ all the nodes are attached both subspaces}

In equation \ref{Eq:D1D2RefVect} $D_{1}+D_{2}$ becomes \begin{eqnarray}
D_{1}+D_{2} & = & -\frac{2}{n}\int d\mathbf{x}\Pr\left(\mathbf{x}\right)\sum_{y=1}^{M}\Pr\left(y|\mathbf{x}\right)\nonumber \\
 &  & \ \times\left\Vert \left(\int d\mathbf{x}_{1}\Pr\left(\mathbf{x}_{1}|y\right)\mathbf{x}_{1},\int d\mathbf{x}_{2}\Pr\left(\mathbf{x}_{2}|y\right)\mathbf{x}_{2}\right)\right\Vert ^{2}\nonumber \\
 &  & \ -\frac{2\left(n-1\right)}{n}\int d\mathbf{x}\Pr\left(\mathbf{x}\right)\nonumber \\
 &  & \ \times\left\Vert \sum_{y=1}^{M}\Pr\left(y|\mathbf{x}\right)\left(\int d\mathbf{x}_{1}\Pr\left(\mathbf{x}_{1}|y\right)\mathbf{x}_{1},\int d\mathbf{x}_{2}\Pr\left(\mathbf{x}_{2}|y\right)\mathbf{x}_{2}\right)\right\Vert ^{2}\nonumber \\
\  & = & -\left(\frac{2}{n}+\frac{2\left(n-1\right)}{n}\right)2R_{\sqrt{M}}\nonumber \\
\  & = & -4R_{\sqrt{M}}\end{eqnarray}

\subsection{Type 3 Optimum:\ half the nodes are attached one subspace and half
are attached to the other}

In equation \ref{Eq:D1D2RefVect} $D_{1}+D_{2}$ becomes \begin{eqnarray}
D_{1}+D_{2} & = & -\frac{2}{n}\left(\frac{2n}{n+1}\right)^{2}\int d\mathbf{x}\Pr\left(\mathbf{x}\right)\nonumber \\
 &  & \ \times\left(\begin{array}{c}
\sum_{y=1}^{\frac{M}{2}}\Pr\left(y|\mathbf{x}\right)\left\Vert \left(\int d\mathbf{x}_{1}\Pr\left(\mathbf{x}_{1}|y\right)\mathbf{x}_{1},0\right)\right\Vert ^{2}\\
+\sum_{y=\frac{M}{2}+1}^{M}\Pr\left(y|\mathbf{x}\right)\left\Vert \left(0,\int d\mathbf{x}_{2}\Pr\left(\mathbf{x}_{2}|y\right)\mathbf{x}_{2}\right)\right\Vert ^{2}\end{array}\right)\nonumber \\
 &  & \ -\frac{2\left(n-1\right)}{n}\left(\frac{2n}{n+1}\right)^{2}\int d\mathbf{x}\Pr\left(\mathbf{x}\right)\nonumber \\
 &  & \ \times\left\Vert \left(\begin{array}{c}
\sum_{y=1}^{\frac{M}{2}}\Pr\left(y|\mathbf{x}\right)\int d\mathbf{x}_{1}\Pr\left(\mathbf{x}_{1}|y\right)\mathbf{x}_{1},\\
\sum_{y=\frac{M}{2}+1}^{M}\Pr\left(y|\mathbf{x}\right)\int d\mathbf{x}_{2}\Pr\left(\mathbf{x}_{2}|y\right)\mathbf{x}_{2}\end{array}\right)\right\Vert ^{2}\nonumber \\
\  & = & -\left(\frac{2n}{n+1}\right)^{2}\left(2\frac{1}{2}\frac{2}{n}+2\left(\frac{1}{2}\right)^{2}\frac{2\left(n-1\right)}{n}\right)R_{\frac{M}{2}}\nonumber \\
\  & = & -\frac{4n}{n+1}R_{\frac{M}{2}}\end{eqnarray}


\begin{thebibliography}{2}
\bibitem{Ref:LuttrellSOM} Luttrell S P, 1994, A Bayesian analysis
of self-organising maps, \textit{Neural Computation}, \textbf{6},
767-794.

\bibitem{Ref:LuttrellPMD} Luttrell S P, 1994, Partitioned mixture
distribution: an adaptive Bayesian network for low-level image processing,
\textit{IEE Proc. Vision Image Signal Processing}, \textbf{141}, 251-260. 
\end{thebibliography}
\end{document}